\documentclass[10pt,twocolumn,letterpaper]{article}

\usepackage[pagenumbers]{arxiv} 

\definecolor{cvprblue}{rgb}{0.21,0.49,0.74}
\usepackage[pagebackref,breaklinks,colorlinks,allcolors=cvprblue]{hyperref}

\usepackage{overpic}
\usepackage{booktabs}
\usepackage{xcolor}
\usepackage{graphicx}
\usepackage{caption}
\usepackage{subcaption}
\usepackage{lipsum} 
\usepackage{multirow}
\usepackage{booktabs} 
\usepackage{amsfonts}
\usepackage{amsmath}
\usepackage{placeins} 
\usepackage{cuted} 
\usepackage{float}  
\usepackage{xcolor}
\usepackage{amssymb}
\usepackage{pifont}
\usepackage{algpseudocode}
\usepackage[ruled,vlined]{algorithm2e}
\SetKwInOut{Input}{Input}
\SetKwInOut{Output}{Output}
\usepackage[table,xcdraw]{xcolor}
\usepackage{bm}

\def\paperID{3582} 
\def\confName{CVPR}
\def\confYear{2026}

\title{\name
}

\author{
{Dianbing Xi$^{1,2,\ast}$, 
Jiepeng Wang$^{2,\ast,\ddagger}$,
Yuanzhi Liang$^{2}$,
Xi Qiu$^{2}$,
Jialun Liu$^{2}$,
Hao Pan$^{3}$
Yuchi Huo$^{1}$,} \\
{Rui Wang$^{1,\dagger}$,
Haibin Huang$^{2}$,
Chi Zhang$^{2}$,
Xuelong Li$^{2,\dagger}$} \\
$^1$ State Key Laboratory of CAD\&CG, Zhejiang University \\
$^2$ Institute of Artificial Intelligence, China Telecom (TeleAI) \\
$^3$ Tsinghua University \\
\url{https://tele-ai.github.io/CtrlVDiff/}
}

\newcommand{\name}{{\emph{CtrlVDiff}: Controllable Video Generation via Unified Multimodal Video Diffusion
}}
\newcommand{\abbname}{\emph{CtrlVDiff}}
\newcommand{\mydata}{\emph{MMVideo}}
\newcommand{\before}{\emph{OmniVDiff}}
\newcommand{\hmcsall}{\emph{Hybrid Modality Control Strategy (HMCS)}}
\newcommand{\hmcs}{\emph{HMCS}}

\newcommand{\cmark}{\textcolor{mygreen}{\checkmark}}
\newcommand{\xmark}{\textcolor{myred}{\ding{55}}}
\newcommand{\umark}{\textcolor{myorange}{\checkmark}}

\definecolor{mygreen}{rgb}{0.0, 0.6, 0.0}        
\definecolor{myred}{rgb}{0.8, 0.1, 0.1}
\definecolor{myorange}{rgb}{1.0, 0.6, 0.0}

\usepackage{xcolor}
\definecolor{c1}{RGB}{31,119,180}
\definecolor{c2}{RGB}{255,127,14}
\definecolor{c3}{RGB}{44,160,44}
\definecolor{c4}{RGB}{214,39,40}
\definecolor{c5}{RGB}{148,103,189}
\definecolor{c6}{RGB}{140,86,75}

\begin{document}
\maketitle
\renewcommand{\thefootnote}{\fnsymbol{footnote}}
\footnotetext[1]{Equal contribution, $^\dagger$Corresponding author, $^\ddagger$Project lead.}
\begin{abstract}
We tackle the dual challenges of video understanding and controllable video generation within a unified diffusion framework. Our key insights are two-fold: geometry-only cues (e.g., depth, edges) are insufficient: they specify layout but under-constrain appearance, materials, and illumination, limiting physically meaningful edits such as relighting or material swaps and often causing temporal drift. Enriching the model with additional graphics-based modalities (intrinsics and semantics) provides complementary constraints that both disambiguate understanding and enable precise, predictable control during generation.

However, building a single model that uses many heterogeneous cues introduces two core difficulties. Architecturally, the model must accept any subset of modalities, remain robust to missing inputs, and inject control signals without sacrificing temporal consistency. Data-wise, training demands large-scale, temporally aligned supervision that ties real videos to per-pixel multimodal annotations.

We then propose \abbname{}, a unified diffusion model trained with a Hybrid Modality Control Strategy (HMCS) that routes and fuses features from depth, normals, segmentation, edges, and graphics-based intrinsics (albedo, roughness, metallic), and re-renders videos from any chosen subset with strong temporal coherence. To enable this, we build MMVideo, a hybrid real-and-synthetic dataset aligned across modalities and captions. Across understanding and generation benchmarks, \abbname{} delivers superior controllability and fidelity, enabling layer-wise edits (relighting, material adjustment, object insertion) and surpassing state-of-the-art baselines while remaining robust when some modalities are unavailable.
\end{abstract}    
\section{Introduction}
\label{sec:intro}
The pursuit of generative models that can synthesize temporally coherent, semantically meaningful, and user-controllable video content represents a critical frontier in artificial intelligence~\cite{alhaija2025cosmos, wang2024dust3rgeometric3dvision,leroy2024groundingimagematching3d,wang20243dreconstructionspatialmemory,wang2025continuous3dperceptionmodel}. Controllable video generation bridges the gap between high-level intent—expressed through text, sketches, trajectories, or structural priors—and dynamic visual realization, enabling precise manipulation of motion, appearance, and scene composition over time.  Multimodal video models aim to learn structured, predictive scene representations that support downstream reasoning, planning, and control~\cite{VideoJAM,xdb2025OmniVDiff,huang2025voyagerlongrangeworldconsistentvideo,alhaija2025cosmos,DiffusionRenderer}. By integrating diverse signals—depth, semantics, and actions—into a unified generative framework, they can simulate plausible futures, infer missing information, and make decisions under uncertainty.

\begin{figure}[htbp]
  \centering
  \includegraphics[width=\columnwidth]{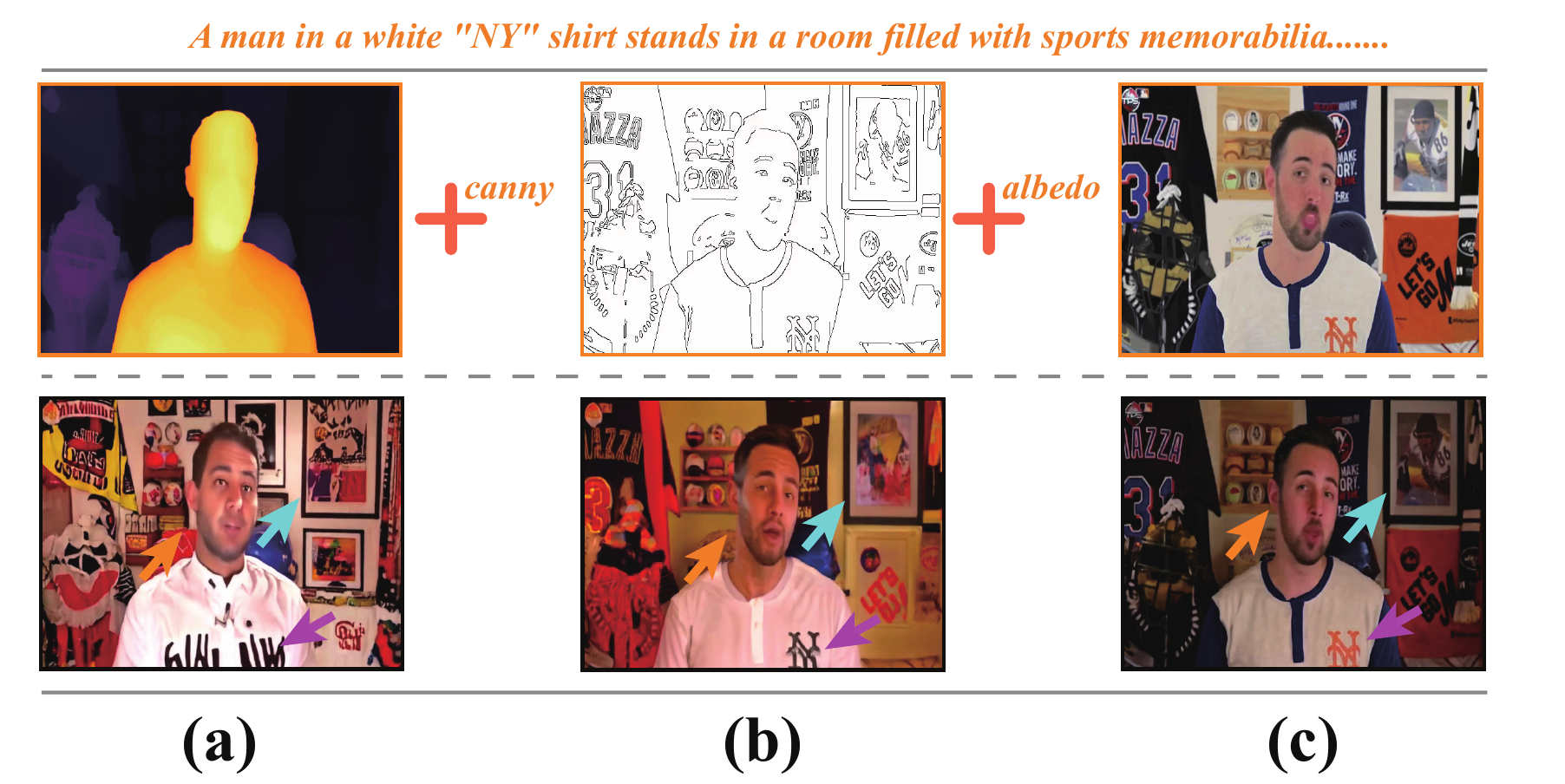}
    \caption{
    \textbf{Impact of different modality combinations on video generation.} 
    Visualization of \abbname{} multimodal generation results. 
    (a) Using only \textit{depth} fails to control facial details and text regions described in the prompt. 
    (b) Combining \textit{depth} and \textit{canny} enables control over facial features (\textcolor[HTML]{f7931e}{\boldmath$\rightarrow$}) and partial text regions 
    (\textcolor[HTML]{ff00ff}{\boldmath$\rightarrow$}).
    (c) Adding \textit{albedo} further refines color and texture control, especially for the background mural (\textcolor[HTML]{00FFFF}{\boldmath$\rightarrow$}).
    }
  \label{fig:intro_analysis}
\end{figure}

However, most existing methods still offer limited controllability. As illustrated in Fig.\ref{fig:intro_analysis}(a), conditioning solely on \textit{depth} constrains layout but leaves appearance under-specified, so prompts alone cannot enforce fine facial attributes or material details, leading to uncontrolled generation. Even systems with “multimodal” control, such as {COSMOS}\cite{alhaija2025cosmos}, largely emphasize geometry/layout cues (e.g., \textit{depth}, \textit{segmentation}, \textit{canny}) and often rely on external expert estimators to obtain control signals. As shown in Fig.~\ref{fig:intro_analysis}(b), this focus omits appearance-related priors (color, texture, material), yielding variations in color, text, and pattern and, ultimately, \emph{uncontrolled video appearance}. Moreover, dependence on external experts introduces domain shift, latency, and error propagation.

Parallel progress on intrinsic-guided diffusion has shown that conditioning on graphics-grounded layers—\textit{albedo}, \textit{normal}, \textit{roughness}, \textit{metallic}—enables photorealistic synthesis with faithful illumination and materials~\cite{huang2025x2videoadaptingdiffusionmodels,liang2025diffusionrenderer,chen2025unirendererunifyingrenderinginverse,Zeng_2024}. Such conditioning lets models reason about underlying physical properties rather than reproducing only coarse structure (see Fig.~\ref{fig:intro_analysis}(c): albedo provides precise control over color and fine texture). However, most AI renderers treat forward rendering (video generation) and inverse rendering (video understanding) as separate problems, with distinct architectures and training pipelines. Inverse pipelines typically recover multiple intrinsic layers (e.g., \textit{depth}, \textit{normals}, \textit{albedo}, \textit{roughness}, \textit{metallic}, \textit{segmentation}), but many predict only \emph{one} layer per pass; obtaining a full stack requires multiple passes, which is computationally costly and prone to cross-layer and temporal inconsistencies. A unified framework that \emph{jointly} learns these layers and supports any-subset control would improve efficiency, coherence, and generalization.

Towards this end, we propose \textbf{\abbname{}}, a controllable video generation framework built on \emph{unified} multimodal video diffusion and supports both \emph{video generation (forward rendering)} and \emph{video understanding (inverse rendering)} within a single model (Fig.~\ref{fig:method_overview}). \abbname{} accepts an arbitrary subset of modalities for conditioning and predicts a temporally consistent stack of outputs, enabling re-rendering with precise, predictable control as well as intrinsic/semantic estimation from rgb. Two key components make this possible.  
First, we design a \textbf{\hmcsall}, 
which stochastically selects conditioning and target modalities, enabling fine-grained controllability while alleviating the optimization difficulty that arises when training across many modalities. 
Under this strategy, the modalities are randomly sampled and categorized as \textit{condition}, \textit{none}, or \textit{noisy} according to predefined probabilities, 
thus improving the robustness of the training and allowing flexible controllable video generation under combinations of arbitrary modality. Second, to overcome the data scarcity inherent in unified multimodal training,  we construct a large-scale unified multimodal dataset, \textbf{\mydata{}}, 
covering \textit{caption} and eight visual modalities: \textit{rgb}, \textit{depth}, \textit{normal}, \textit{albedo}, \textit{roughness}, \textit{metallic}, \textit{segmentation}, and \textit{canny}. 
The dataset combines real and synthetic sources, featuring diverse scenarios like indoor/outdoor scenes, human-centric videos, and object-level captures. This multimodal mix significantly boosts model generalization and cross-domain alignment.

To comprehensively evaluate the effectiveness of our method, 
we conduct experiments across five key dimensions — \textit{depth estimation}, \textit{segmentation estimation}, 
\textit{normal estimation}, \textit{material estimation}, and \textit{video generation}. 
Our method achieves state-of-the-art performance in all these aspects. 
In addition, it supports a variety of high-quality applications, including 
{scene relighting} (Figure~\ref{fig:teaser}(c)), 
{material editing} (Figure~\ref{fig:teaser}(d)), 
and {object insertion} (Figure~\ref{fig:teaser}(e)).

\noindent
{In summary, our main contributions are as follows:}
\begin{enumerate}
    \item We propose {\abbname{}}, a controllable video generation framework based on unified multimodal diffusion, 
    which, to our knowledge, is the first to jointly support both {video generation (forward rendering)} 
    and {video understanding (inverse rendering)} in a single model.

    \item We introduce a {\hmcsall}, 
        which enables flexible and controllable video generation with arbitrary mode combinations, 
        improving stability of training and convergence speed.

    \item We build a large-scale dataset, {\mydata{}}, 
    covering diverse visual modalities from real and synthetic sources, 
    alleviating data scarcity in multimodal learning and enhancing cross-modality generalization.

    \item Extensive experiments show that {\abbname{}} outperforms existing approaches 
    in both quantitative and qualitative evaluations. 
    For video understanding, it achieves comparable or superior performance to expert models trained for {single-modality estimation}.
\end{enumerate}

\section{Related Works}
\label{sec:related_works}

\subsection{Multimodal Video Generation}
Video diffusion models (VDMs)~\cite{brooks2024video,zhu2024compositional,kong2024hunyuanvideo,zheng2024open,polyak2025moviegencastmedia,wan2025} 
achieve realistic and temporally consistent video synthesis. 
Recent studies on controllable video generation aim for fine-grained control over synthesized content. 
To enhance controllability, many approaches incorporate multimodal signals (e.g., \textit{depth}, \textit{edges}, \textit{segmentation}, \textit{3D cues}) as conditions~\cite{zhang2023addingconditionalcontroltexttoimage,esser2023runwaygen1,zhai2024idol,VideoJAM}. 
ControlNet~\cite{zhang2023addingconditionalcontroltexttoimage} augments pre-trained diffusion models with lightweight zero-initialized control branches, 
while Gen-1~\cite{esser2023runwaygen1} decouples structure from content. 
Recent works further unify multimodal generation, such as IDOL~\cite{zhai2024idol}, which jointly generates rgb and depth, 
and VideoJAM~\cite{VideoJAM}, which extends this to rgb and motion.  

Despite recent progress, existing methods suffer from two critical drawbacks:  
(1) {Lack of modality-agnostic control}:  Each new control signal often demands dedicated fine-tuning of the generative model, leading to fragmented workflows and limited cross-modal transferability;  
(2) {External dependency}: most methods depend on conditioning signals extracted by specialized models, limiting flexibility and generalization.

\subsection{Multimodal Video Understanding}
Video understanding, traditionally centered on discriminative tasks such as classification, detection, and segmentation, 
is being redefined by generative modeling. 
A central goal of generative visual understanding is to model fundamental geometric modalities from 2D images and videos, 
including \textit{depth}, \textit{normals}, and \textit{segmentation maps}. 
Recent studies reformulate classical perception tasks as conditional generation problems, 
leveraging priors learned by large-scale diffusion models~\cite{ke2023repurposing,ye2024stablenormal,ke2025marigold,bin2025normalcrafterlearningtemporallyconsistent}. 
NormalCrafter~\cite{bin2025normalcrafterlearningtemporallyconsistent} extends this idea by adapting video diffusion architectures 
to generate temporally consistent sequences of normals.  

Building on single-modality generation, an emerging trend is the joint synthesis of multiple coupled 3D and spatio-temporal modalities 
to build holistic and coherent scene representations~\cite{fu2024geowizard,he2024lotus,xu2025geometrycrafter,jiang2025geo4d,zhao2025diception,krishnan2025orchid,byung2025jointdit}.  
Among these approaches, JointDiT~\cite{byung2025jointdit} leverages a diffusion transformer to model the joint distribution of RGB and depth signals. 
It enables unconditional image synthesis, depth estimation, and depth-conditioned generation through adaptive weighting strategies and unbalanced timestep sampling.
This line of research underscores the importance of multimodal integration for advancing scene understanding. However, most existing methods focus on a limited set of modality pairs. 
Extending to broader modalities and enabling flexible conditioning remain key yet underexplored challenges.

\subsection{Unified  Multimodal Video Model.}
In recent years, unified multimodal video modeling has emerged as a prominent trend, 
aiming to integrate diverse vision tasks within a single end-to-end framework. 
In the image domain, many studies have explored generation and understanding using unified diffusion frameworks~\cite{li2024unicon,wang2025mmgen,byung2025jointdit,wu2025omnigen2}. 
For example, {MMGen}~\cite{wang2025mmgen} unifies multimodal generation and understanding within a single transformer, supporting category-conditioned generation and controllable synthesis. 
Building on these advances, recent efforts have extended such ideas to the video domain to capture temporal dynamics and maintain cross-modal consistency~\cite{VideoJAM,aether,xdb2025OmniVDiff,yang2025omnicamunifiedmultimodalvideo,DiffusionRenderer}. 
AETHER~\cite{aether} post-trains a video diffusion model on synthetic 4D RGB-D data and camera trajectories, enabling zero-shot 4D reconstruction and goal-driven visual planning. 
OmniVDiff~\cite{xdb2025OmniVDiff} models the joint distribution of rgb, depth, segmentation, and edges via a shared 3D-VAE with adaptive modality embeddings, supporting video generation and video understanding. 

These works highlight the value of controllable video representations. 
Although unified multimodal video models have made initial progress, they mainly focus on architectural unification for efficiency. 
However, these approaches still lack fine-grained controllability in video generation, and precise control remains challenging. 
In this paper, we address these problems through unified multimodal video diffusion, which preserves model efficiency while achieving high controllability in video generation, thus overcoming the limitations of previous methods.

\section{Method}
\label{sec:method}
\begin{figure*}[t]
    \includegraphics[width=1.0\linewidth]{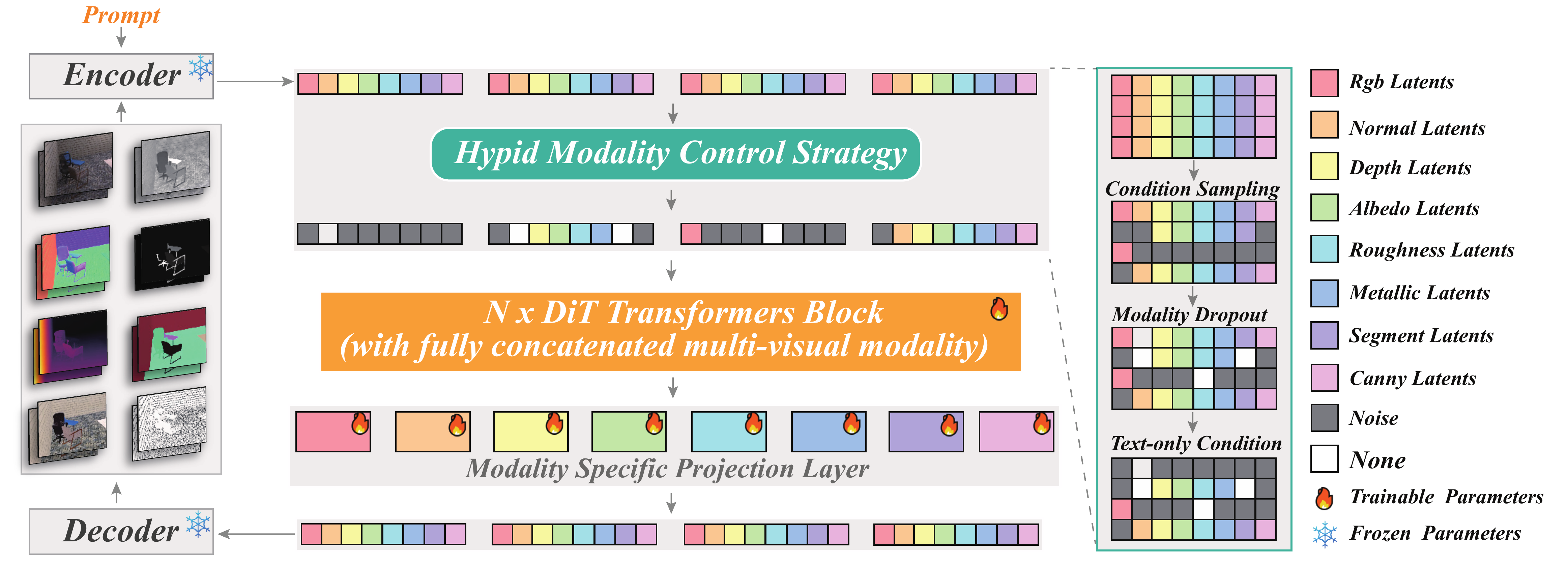}
    \caption{
    Framework overview of \textbf{ \abbname{}}. 
    Given a video with eight paired modalities, 
    we first encode all modalities into latent representations using a pretrained shared 3D-VAE encoder. 
    For each sample within a batch, its latent features are concatenated along the channel dimension. 
    Subsequently, we apply the {\hmcs{}} to each batch (as illustrated in the \textcolor[HTML]{00a99d}{box} on the right), 
    which enables robust handling of all possible modality combinations. 
    The outputs of the Diffusion Transformer are then processed through {modality specific projection layers}, 
    where each modality is assigned an independent projection head to encourage effective modality disentanglement. 
    }
    \vspace{-0.4cm}
   \label{fig:method_overview}
\end{figure*}

In this section, we present {\abbname{}}, a controllable video generation framework that jointly models four categories of scene properties—\textit{geometry} (depth, normal), \textit{appearance} (albedo, roughness, metallic), \textit{semantics} (segmentation), and \textit{structure} (canny)—to achieve precise and interpretable control over video generation. 
We first introduce the {\hmcsall{}} (Section~\ref{sec:hmcs}), which enables arbitrary modality combinations while maintaining training stability. 
Next, we describe our {data annotation pipeline} (Section~\ref{sec:data_pipeline}) for large-scale, fine-grained supervision across diverse video modalities. 
Finally, we outline the overall {training paradigm} in Section~\ref{sec:training_paradigm}. 
An overview of the framework is shown in Figure~\ref{fig:method_overview}.

\subsection{Hypid Modality Control Strategy}
\label{sec:hmcs}

OmniVDiff~\cite{xdb2025OmniVDiff} demonstrates the capability to synthesize and comprehend diverse video modalities within a unified diffusion framework by representing all modalities in a shared color space and learning their joint distribution. 
Following this foundation, we adopt CogVideoX~\cite{yang2024cogvideox} as our base video model.
We propose {\hmcsall}, which supports flexible modality combinations and enhances training robustness across a broader set of modalities.

Our strategy dynamically determines which modalities serve as conditions or generation targets, improving generalization and preventing overfitting to fixed configurations. 
Formally, given $N$ modalities $\{M_1, M_2, \dots, M_N\}$, \hmcs{} operates in four stages:
\begin{enumerate}
    \item \textbf{Conditional Sampling.} 
    Randomly select $k \in [1, N-1]$ modalities as the conditional set $\mathcal{C}$ (\textit{condition}), with the remaining used for generation.
    \item \textbf{Modality Dropout.} 
    Randomly drop $d \in [1, N-1]$ modalities from $\mathcal{M}$ as the drop set $\mathcal{D}$, marking them as \textit{none} to simulate missing modalities.
    \item \textbf{Text-only Condition.} 
    For a subset of samples, replace $\mathcal{C}$ with $\{\text{text}\}$ to ensure text-driven generation capability, 
    and mark the corresponding $\mathcal{M}$ as \emph{noise}.
    \item \textbf{Generation Target Selection.} 
    We define $\mathcal{G} = \mathcal{M} \setminus (\mathcal{C} \cup \mathcal{D})$.
    For each $M_i \in \mathcal{G}$, we apply a Gaussian perturbation (denoted as \textit{noise}).
    \begin{equation}
        \tilde{x}_i = \sqrt{\alpha_t} x_i + \sqrt{1 - \alpha_t}\,\epsilon, \quad \epsilon \sim \mathcal{N}(0, \mathbf{I}).
    \end{equation}
\end{enumerate}
This stochastic control mechanism encourages modality-agnostic representations and flexible adaptation to varying modality configurations during training and inference.
The detailed algorithm is presented in Algorithm~\ref{algorithm:hmcs}.
\begin{algorithm}
\caption{\textbf{Hybrid Modality Control Strategy (HMCS)}}
\label{alg:hmcs}
\Input{Modality set $\mathcal{M}=\{M_1,\dots,M_N\}$; text prompt $T$}
\Output{Conditional set $\mathcal{C}$; generation set $\mathcal{G}$}

\BlankLine
\textbf{Initialize} $\mathcal{D}\leftarrow\varnothing$; \Comment{dropped (missing) modalities, mark as none}

\BlankLine
\textbf{Conditional sampling:} sample $k \sim \mathcal{U}\{1,\dots,N-1\}$; \;
$\mathcal{C}\leftarrow \textsf{Sample}(\mathcal{M},k)$; mark each $M_i\in\mathcal{C}$ as \textit{condition};\;

\BlankLine
\textbf{Modality dropout:} sample $d \sim \mathcal{U}\{1,\dots,N-1\}$; \;
$\mathcal{D}\leftarrow \textsf{Sample}(\mathcal{M},d)$; mark each $M_i\in\mathcal{D}$ as \textit{None};\;

\BlankLine
\textbf{Text-only conditioning (optional):} 
With probability $p_t$, we set $\mathcal{C} \leftarrow \{T\}$, 
and mark all non-\textit{None} modalities as \textit{noise}.

\BlankLine
\textbf{Generation target selection:}
$\mathcal{G}\leftarrow \mathcal{M}\setminus(\mathcal{C}\cup\mathcal{D})$;\;
\ForEach{$M_i \in \mathcal{G}$}{
    Add Gaussian noise for diffusion training:
    $\tilde{x}_i \leftarrow \sqrt{\alpha_t}\,x_i + \sqrt{1-\alpha_t}\,\epsilon,\ \epsilon\sim\mathcal{N}(0,\mathbf{I})$; \;
    mark $M_i$ as \textit{noisy}; \;
}
\Return{$\mathcal{C},\ \mathcal{G}$}; \;
\label{algorithm:hmcs}
\end{algorithm}

\subsection{Data Annotation Pipeline}
\label{sec:data_pipeline}
\begin{table}[ht]
\centering
\renewcommand{\arraystretch}{1.1}
\setlength{\tabcolsep}{5pt}
\caption{Comparison of interior scene datasets across multiple modalities. Each data channel is categorized as ``available'' (\cmark), ``unavailable'' (\xmark), or ``available but unreliable'' (\umark).}
\resizebox{1.0\linewidth}{!}{     
\begin{tabular}{lcccccccc}
\toprule
 & Size & Depth & Normal & Albedo  & Roughness & Metallic & Segment & Smooth \\
\midrule
HyperSim    &50k  & \cmark  & \cmark & \umark & \xmark & \xmark & \cmark & \xmark \\
InteriorVerse &73k & \cmark & \cmark & \cmark     & \umark & \umark & \xmark & \xmark \\
InteriorNet   &20M & \cmark & \cmark & \cmark & \xmark & \xmark & \cmark & \umark \\
\textbf{MMVideo(Ours)} & 350k & \cmark & \cmark & \cmark & \cmark & \cmark & \cmark & \cmark \\
\bottomrule
\end{tabular}
} 
\label{tab:dataset_compare}
\end{table}

Open-source datasets~\cite{zhu2022learningbasedinverserenderingcomplex,roberts2021hypersimphotorealisticsyntheticdataset,li2018interiornetmegascalemultisensorphotorealistic,li2023matrixcity} 
provide images of indoor and outdoor scenes accompanied by corresponding multimodal data. 
As shown in Table~\ref{tab:dataset_compare}, these datasets face key challenges: 
inaccurate intrinsic signals, unsmooth camera motions, and exclusive reliance on synthetic content, 
all of which limit their utility in realistic video applications.

To address these limitations, we present \textbf{MMVideo}, 
a novel video dataset that provides a comprehensive and reliable suite of multimodal annotations, 
covering both real-world and synthetic environments.

\textbf{Synthetic data.} 
We adopt two primary generation pipelines. 
The first pipeline uses 3D-Front~\cite{fu20213dfront3dfurnishedrooms} indoor layouts as geometric foundations. 
Since 3D-Front lacks physically based rendering (PBR) materials, 
we randomly assign 1,824 high-quality PBR materials from ambientCG~\cite{ambientcg} according to object semantic labels. 
The second pipeline builds indoor scenes from the ABO~\cite{collins2022abodatasetbenchmarksrealworld} dataset, 
which already provides native PBR materials. 
In total, we generate approximately 100K synthetic video clips.

\textbf{Real data.} 
The real portion of MMVideo comprises 200K video clips from Koala-36M~\cite{wang2024koala}, 
augmented with multimodal annotations inferred by expert models and supplemented by open-source datasets that originally lacked certain modalities. 

Overall, {MMVideo} contains 350K video clips at 16 fps with 49 frames each. 
It spans a broad spectrum of real-world and synthetic scenarios—covering indoor and outdoor environments, 
dynamic and static scenes, and diverse subjects such as humans, animals, and complex objects—demonstrating strong diversity and generalization potential.

\subsection{Training Paradigm}
\label{sec:training_paradigm}

Our overall training framework is designed as a three-stage process, enabling progressive optimization and refinement.

\textbf{Stage I.} We first train the model for text-conditioned video generation, enabling direct synthesis of multimodal videos from textual prompts.
\textbf{Stage II.} We incorporate the \hmcs{} module to achieve controllable video generation. This strategy dynamically assigns each modality as either a conditioning input or a generation target, supporting controllable synthesis under arbitrary modality combinations.

Through these two stages, the model learns unified generation and understanding capabilities. 
However, due to the domain gap between synthetic and real data, its understanding ability on real videos remains limited. 
Inspired by SAM2~\cite{sam2}, we introduce a \textbf{self-augmentation phase} as the \textbf{Stage III} to enhance performance on real-world data.

Specifically, we first train a model on a small synthetic subset to improve understanding accuracy, 
then use it to annotate 40K real video samples. 
After automatic validation and manual refinement, we curate 20K high-quality samples. 
Finally, we jointly fine-tune on synthetic and curated real data under the \hmcs{} framework for 1K iterations, 
further improving video generation and understanding.

To jointly train video {generation} and {understanding} across modalities, 
we adopt a modality-wise diffusion objective that maintains sample quality while being agnostic to conditioning configuration. 
Each modality is optimized independently with a denoising loss; labeled as \textit{condition} or \textit{none}, they are excluded from reconstruction.
Let \(\mathrm{Cond}\) denote condition modalities and \(m\) index a modality with embedding \(e_m\). 
The overall objective is:

\begin{equation}
\mathcal{L}
= \sum_{m} \mathbf{1}[m \notin \mathrm{Cond}]
\mathbb{E}_{\tilde{\mathbf{x}}_{m,t},t,\epsilon}
\Big[
\|\epsilon - \epsilon_{\theta}(\tilde{\mathbf{x}}_{m,t}, t, e_m)\|_2^2
\Big],
\end{equation}
where \(\epsilon_{\theta}\) is the noise-prediction network and \(\epsilon\) the Gaussian noise. 
This {masked supervision} allows dynamic role reassignment among modalities, 
enabling seamless transitions between conditioning and generation without retraining.
\section{Experiments}

To comprehensively evaluate both video generation and understanding, we follow the evaluation protocol of~\cite{xdb2025OmniVDiff} and report video generation results on {VBench}~\cite{huang2023vbench}. Detailed evaluation protocols for each video understanding modality are provided in the Appendix.

\subsection{Video Understanding}

We comprehensively evaluate the video understanding component to assess how accurately the model predicts each modality serving as a conditioning signal for generation. 
For material estimation, please refer to the Appendix.

\begin{figure}[htbp]
    \centering
    \includegraphics[width=1.0\linewidth]{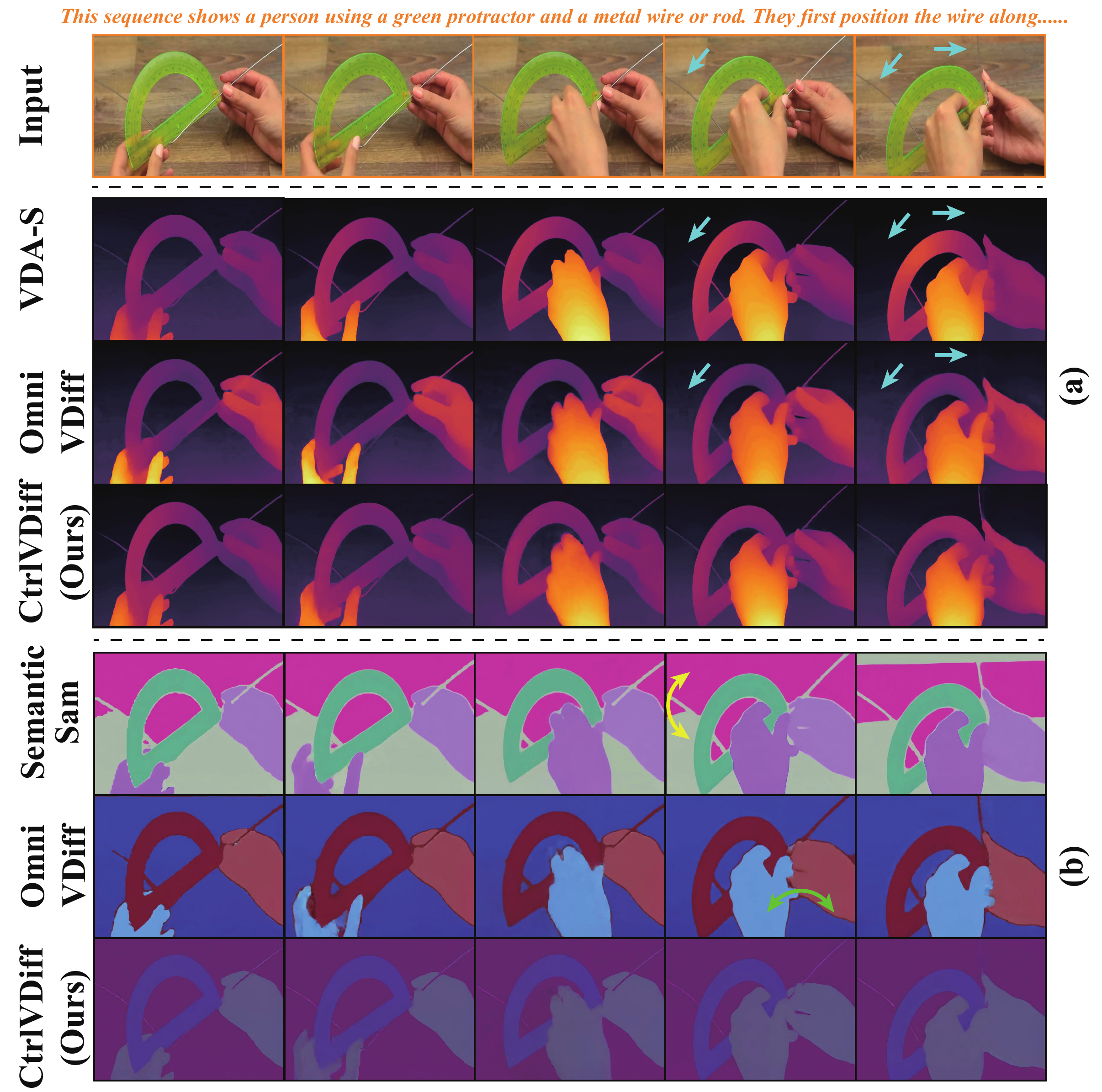}
    \caption{
    \textbf{Qualitative comparison of video depth and segmentation estimation.} 
    (a) \textbf{Video Depth Estimation:} \textit{VDA-S} denotes the \textbf{Video Depth Anything} expert model with a ViT-Small backbone. 
    The \textcolor[HTML]{00FFFF}{\boldmath$\rightarrow$} highlight that \abbname{} consistently predicts accurate depth for fine structures such as thin wires. 
    (b) \textbf{Video Segmentation Estimation:} 
    The \textcolor[HTML]{fcee21}{\boldmath$\rightarrow$} indicate regions that are incorrectly segmented into multiple classes due to object occlusion, 
    while the \textcolor[HTML]{00ff00}{\boldmath$\rightarrow$} mark ambiguous regions where the segmentation granularity is inconsistent. 
    \abbname{} achieves the best performance across both tasks.
    }
    \vspace{-0.2cm}
    \label{fig:est_depth_seg}
\end{figure}

\textbf{Depth Estimation.} As shown in Table~\ref{tab:est_depth} and Figure~\ref{fig:est_depth_seg} (a), 
\abbname{} achieves state-of-the-art performance among all baselines, 
delivering results comparable to the expert model {VDA-S}. 
Notably, {VDA-S} serves as our expert model and is trained with high-quality ground-truth depth supervision. 
Under our proposed train strategy and the {MMVideo} dataset, 
\abbname{} demonstrates superior capability in estimating the depth of thin and fine-grained structures compared with related baselines.

\textbf{Segment Estimation.} 
As shown in Table~\ref{tab:est_seg} and Figure~\ref{fig:est_depth_seg}(b), \abbname{} achieves performance comparable to expert models. 
In particular, our method effectively avoids {incorrect segmentation into multiple classes caused by object occlusion} 
and mitigates {ambiguous regions} where segmentation granularity is inconsistent. 
More importantly, we observe that \abbname{} produces more accurate segmentation for {thin structures} 
and yields smoother results overall, which can be attributed to our Stage~III refinement.

\begin{table}[t]
\centering
\caption{\textbf{Quantitative comparison: Zero-shot video depth estimation results.} Comparison of performance across representative single-image and video-based depth estimation models. ``VDA-S(e)'' denotes the expert model with a ViT-Small backbone. 
The \textbf{best} and \underline{second-best} results are emphasized for clarity.}
\begin{tabular}{ccc}
\toprule
Method & AbsRel $\downarrow$ & $\delta_1$ $\uparrow$  \\
\midrule
DAv2-L\cite{yang2024depthv2}             & 0.150 & 0.768  \\
NVDS\cite{wang2023neural}               & 0.207 & 0.628  \\
NVDS + DAv2-L      & 0.194 & 0.658  \\
ChoronDepth \cite{shao2024learningtemporallyconsistentvideo}        & 0.199 & 0.665  \\
DepthCrafter\cite{hu2024depthcraftergeneratingconsistentlong}       & 0.169 & 0.730  \\

\bottomrule
VDA-S (e)\cite{video_depth_anything}          & \underline{0.110} & \underline{0.876}  \\
\before~\cite{xdb2025OmniVDiff}   &  0.125 & 0.852  \\

\abbname(Ours) & \textbf{0.105} &  \textbf{0.889} \\

\bottomrule
\end{tabular}
\vspace{-0.3cm}
\label{tab:est_depth}
\end{table}
\begin{table}[h]
    \centering
        \caption{\textbf{Comparison with prior methods on point-based interactions, evaluated on COCO Val2017.} “Max” selects the prediction with the highest confidence score, while “Oracle” uses the one with highest IoU against the target mask. The \textbf{best} and \underline{second-best} results are emphasized for clarity.}
    \resizebox{0.5\textwidth}{!}{ 
    \begin{tabular}{lcc}
        \toprule
        Method & \multicolumn{2}{c}{COCO Val 2017\cite{lin2015microsoftcococommonobjects}} \\
        \cmidrule(lr){2-3}
        & Point (Max) 1-IoU $\uparrow$ & Point (Oracle) 1-IoU $\uparrow$ \\
        \midrule
        SAM (B)\cite{kirillov2023segment}            & 52.1 & 68.2 \\
        SAM (L)\cite{kirillov2023segment}            & 55.7 & 70.5 \\
        Semantic-SAM (T)\cite{li2023semantic}   & 54.5 & 73.8 \\
        Semantic-SAM (L)(e)\cite{li2023semantic}   & \textbf{57.0} & \textbf{74.2} \\
        \before~\cite{xdb2025OmniVDiff} & {56.0} & {73.9}\  \\
        \abbname(Ours) & \textbf{57.0} & \underline{74.1}\  \\
        \bottomrule
    \end{tabular}
    } 

    \label{tab:est_seg}
\end{table}

\begin{figure}
    \centering
    \includegraphics[width=1.0\linewidth]{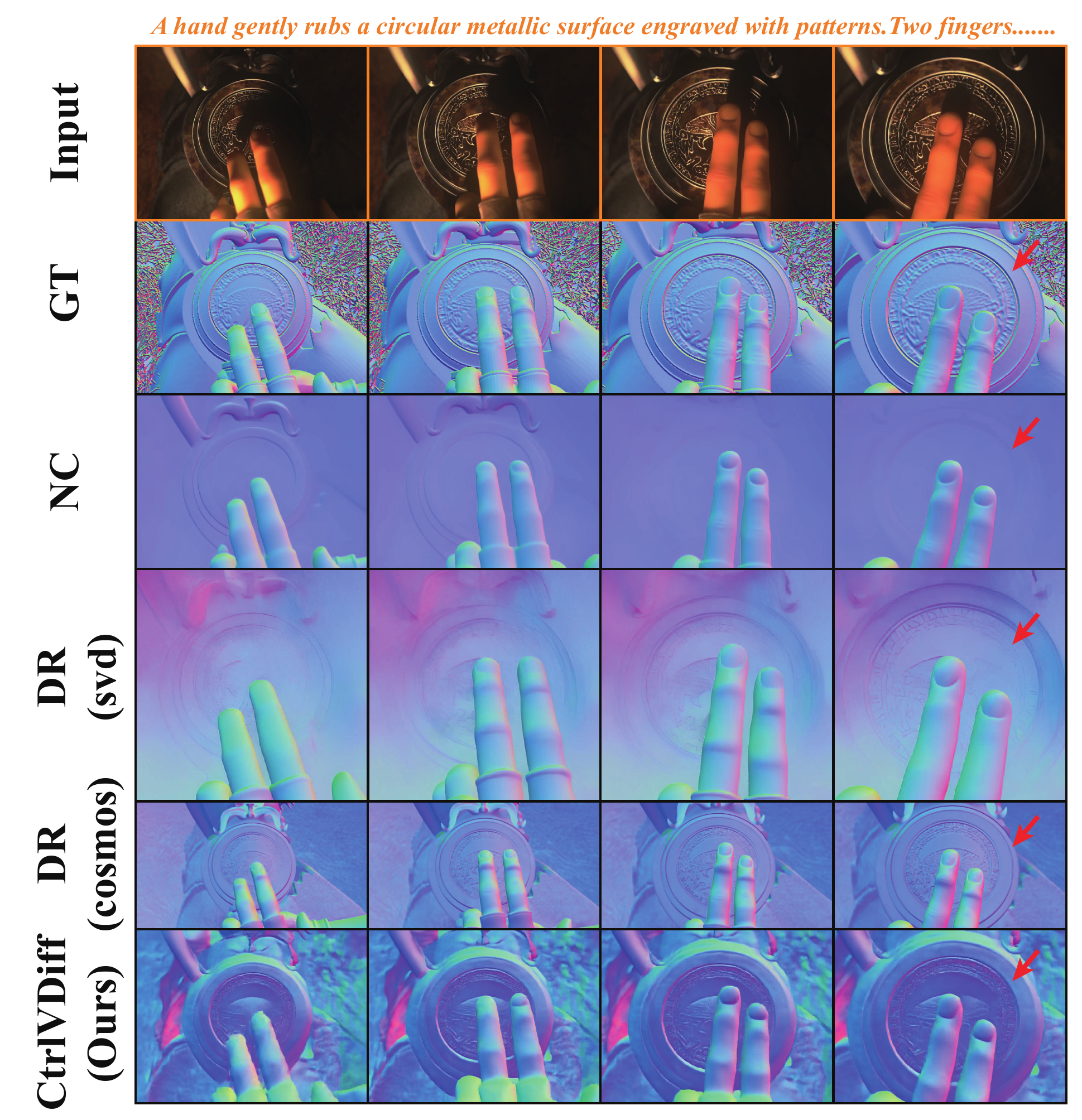}
    \caption{\textbf{Qualitative comparison of video normal estimation.}
NormalCrafter is denoted as NC, and DiffusionRenderer as DR.
Both \abbname{} and DR (Cosmos) demonstrate superior performance in preserving fine details and surface consistency(\textcolor[HTML]{ff1d25}{\boldmath$\rightarrow$}). }
\vspace{-0.4cm}
\label{fig:est_normal}
\end{figure}

\textbf{Normal Estimation.} As shown in Table~\ref{tab:est_normal} and Figure~\ref{fig:est_normal}, 
\abbname{} achieves performance comparable to the expert model {DiffusionRenderer (Cosmos)} and demonstrates even stronger performance on the {ScanNet} dataset. 
We attribute this primarily to inheriting the performance of our expert model. Additionally, through the design of Stage 3, our model surpasses the expert model in terms of overall performance.
Compared with single-modality normal estimation methods such as {NormalCrafter}~\cite{bin2025normalcrafterlearningtemporallyconsistent}, 
\abbname{} also demonstrates competitive performance, especially on the {Sintel} dataset. As illustrated in Figure~\ref{fig:est_normal}, 
NormalCrafter tends to produce overly smooth results in motion-heavy scenes involving human–object interactions, 
whereas our method preserves richer structural details and yields more realistic surface geometry. This advantage primarily stems from our model being trained on a broader range of data, which endows it with stronger generalization capability.

\begin{table*}
    \centering

    \label{tab:est_normal}
    \caption{\textbf{Quantitative evaluation on ScanNet and Sintel video benchmarks (angles in degrees).} 
    Higher is better for thresholds; lower is better for mean, median, and rank. 
    The \textbf{best} and \underline{second-best} results are emphasized for clarity.}
    \resizebox{\textwidth}{!}{
    \begin{tabular}{lcccccc|cccccc}
        \toprule
        & \multicolumn{6}{c}{\textbf{ScanNet}~\cite{dai2017scannetrichlyannotated3dreconstructions} (Video Benchmark)} 
        & \multicolumn{6}{c}{\textbf{Sintel}~\cite{10.1007/978-3-642-33783-3_44} (Video Benchmark)} \\
        \cmidrule(lr){2-7}\cmidrule(lr){8-13}
        \textbf{Method} 
        & mean $\downarrow$ & med $\downarrow$ & $11.25^\circ$ $\uparrow$ & $22.5^\circ$ $\uparrow$ & $30^\circ$ $\uparrow$ & Rank $\downarrow$
        & mean $\downarrow$ & med $\downarrow$ & $11.25^\circ$ $\uparrow$ & $22.5^\circ$ $\uparrow$ & $30^\circ$ $\uparrow$ & Rank $\downarrow$ \\
        \midrule
        DSINE~\cite{bae2024rethinkinginductivebiasessurface}                 & 15.5 & 8.0 & 62.4 & 79.5 & 84.9 & 6.4 & 34.9 & 28.1 & 21.5 & 41.5 & 52.7 & 6.4 \\
        GeoWizard~\cite{fu2024geowizardunleashingdiffusionpriors}         & 18.9 & 13.1 & 41.7 & 57.1 & 83.0 & 9.0 & 37.6 & 32.0 & 11.7 & 32.8 & 46.8 & 8.6 \\
        GenPercept~\cite{xu2024mattersrepurposingdiffusionmodels}       & 14.5 & 7.2 & 66.0 & 81.8 & 86.7 & 3.8 & 34.6 & 26.2 & 18.4 & 43.8 & 55.8 & 5.6 \\
        StableNormal~\cite{ye2024stablenormalreducingdiffusionvariance}   & 15.9 & 10.0 & 57.0 & 81.0 & 87.4 & 6.0 & 34.8 & 32.7 & 17.9 & 36.1 & 46.6 & 8.2 \\
        Lotus-D~\cite{he2025lotusdiffusionbasedvisualfoundation}              & 14.3 & 7.1 & 65.6 & 81.4 & 86.5 & 4.2 & 33.2 & 25.5 & 22.4 & 44.9 & 57.0 & 3.6 \\
        Marigold-E2E-FT~\cite{garcia2025finetuningimageconditionaldiffusionmodels}    & \underline{14.1} & \textbf{6.3} & \underline{67.6} & 81.7 & 86.4 & 3.0 & 33.5 & 27.0 & 21.5 & 43.0 & 54.3 & 5.2 \\
       {NormalCrafter}~\cite{bin2025normalcrafterlearningtemporallyconsistent}                      & \textbf{13.3} & \underline{6.8} & 67.4 & \textbf{82.9} & \textbf{87.9} & \textbf{1.6} 
                                               & \underline{30.7} & {23.9} & \underline{23.5} & {47.5} & {60.1} & \underline{2.4} \\
        {DiffusionRenderer(SVD)}                   & {19.7} & 11.8 & 49.6 & {72.5} & {79.7} & 8.4 & 39.1 & 34.6 & 11.2 & 31.3 &  43.1 & 10 \\
        
        {DiffusionRenderer(cosmos)}                        & {22.2} & 13.8 & {43.2} & {68.8} & {76.9} & 9.6 & 31.0 & \underline{23.8} & \textbf{25.2} & \underline{49.8} & \underline{60.9} & 2.6
                                   \\

        \abbname{}(Ours)              &  {15.4} & {7.5} & \textbf{67.7} & \underline{82.7} & \underline{87.5} & \underline{3} & \textbf{30.6} & \textbf{23.5} & {20.8} & \textbf{49.9}  & \textbf{61.7} & \textbf{1.8}   \\
        
        
        \bottomrule
    \end{tabular}
    }
    \vspace{-0.4cm}
    \label{tab:est_normal}
\end{table*}

\subsection{Controllable Video Generation}

In this section, we conduct a comprehensive evaluation of our framework through both quantitative and qualitative analyses on {single-condition} and {multi-condition} video generation tasks. 
The comparative results for the multi-condition setting are presented in the main paper, while those for the single-condition case are included in the appendix for completeness.

We compare our approach with the most relevant state-of-the-art methods, including {RGBX}~\cite{Zeng_2024} and {DiffusionRenderer}~\cite{DiffusionRenderer} (both \texttt{svd} and \texttt{cosmos} variants). 
During multi-condition generation experiments, all methods are provided with the full set of conditioning modalities to ensure a fair comparison. 
As each method utilizes all available conditions, this setting is equivalent to a video reconstruction task.

\begin{figure}[t]
  \centering
  \includegraphics[width=1.0\linewidth]{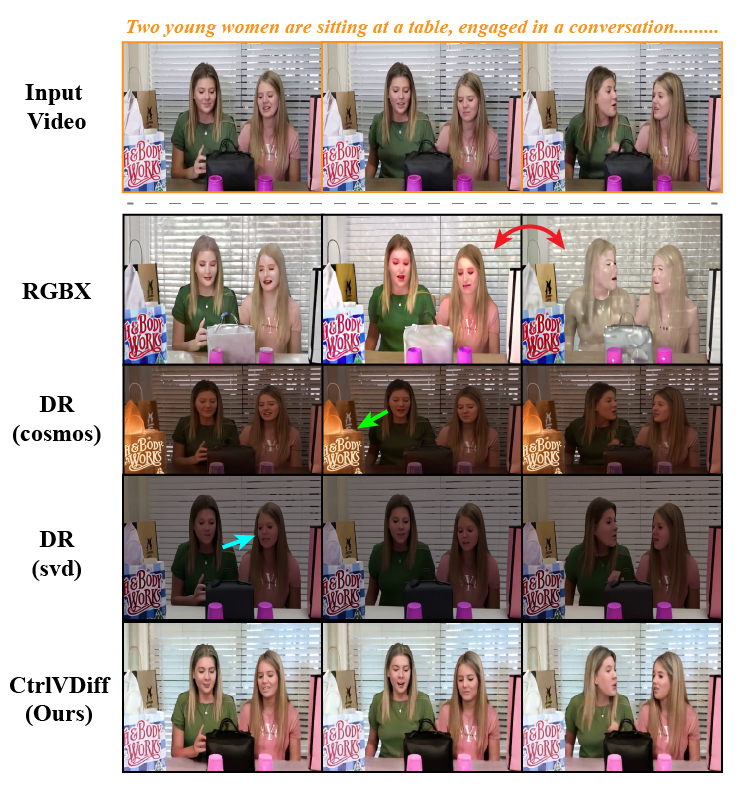}
\caption{ \textbf{Qualitative comparison of multi-condition video generation(video reconstruction).} Compared with our baseline, {RGBX} shows temporal flickering and inconsistencies (\textcolor[HTML]{ff1d25}{\boldmath$\rightarrow$}). {DiffusionRenderer (DR, svd)} exhibits compositing artifacts, especially on faces (\textcolor[HTML]{3fa9f5}{\boldmath$\rightarrow$}). {DR (cosmos)} produces incorrect re-synthesis, such as inaccurate object colors (e.g., the bag; (\textcolor[HTML]{7ac943}{\boldmath$\rightarrow$})). In contrast, our method achieves the most faithful reconstruction using self-decomposed parameters. }
\vspace{-0.5cm}
\label{fig:multi_cond_video_gen}
\end{figure}

As shown in Figure~\ref{fig:multi_cond_video_gen} and Table~\ref{tab:multi_cond_video_gen}, our approach achieves the most faithful reconstruction results compared to the input videos. 
We attribute this improvement primarily to the more accurate estimation of modality representations, which enables better disentanglement and consistency across visual factors. 
Moreover, thanks to our large-scale \mydata{}, the reconstructed videos exhibit higher realism and smoother motion, demonstrating the robustness and generalization ability of our framework.

\begin{table*}[t]
\centering
\caption{\textbf{VBench Evaluation Metrics for Multi Condition Video Generation (Video Reconstruction).} 
For each method, the top-performing result is emphasized in \textbf{bold}, and the second-best performance is marked with an \underline{underline}.}
\resizebox{\textwidth}{!}{
\begin{tabular}{lccccccc}
\toprule
Model & subject consistency & b.g. consistency & motion smoothness & dynamic degree & aesthetic quality & imaging quality & weighted average \\
\midrule
\hspace{1em}RGBX~\cite{Zeng_2024} &{88.12} &{85.49} &{92.51} &{39.21} &{34.79} &{40.04} &{60.09}  \\
\hspace{1em}DR(svd)~\cite{DiffusionRenderer} &{97.77} &{98.16} &{99.23} &{53.75} &{48.73} &{64.63} &{72.56}  \\
\hspace{1em}DR(cosmos)~\cite{DiffusionRenderer} &\underline{98.24} &\textbf{98.49} &\underline{99.45} &\underline{53.92} &\underline{52.41} &\underline{70.91} &\underline{74.41}  \\
\hspace{1em}\abbname{}(Ours) &\textbf{98.73} &\underline{98.46} &\textbf{99.49} &\textbf{54.25} &\textbf{58.72} &\textbf{71.63} &\textbf{75.69}  \\
\bottomrule
\end{tabular}
}
\vspace{-0.4cm}
\label{tab:multi_cond_video_gen}
\end{table*}

\subsection{Ablation study}
\begin{figure}[htbp]
  \centering
  \includegraphics[width=0.45\textwidth]{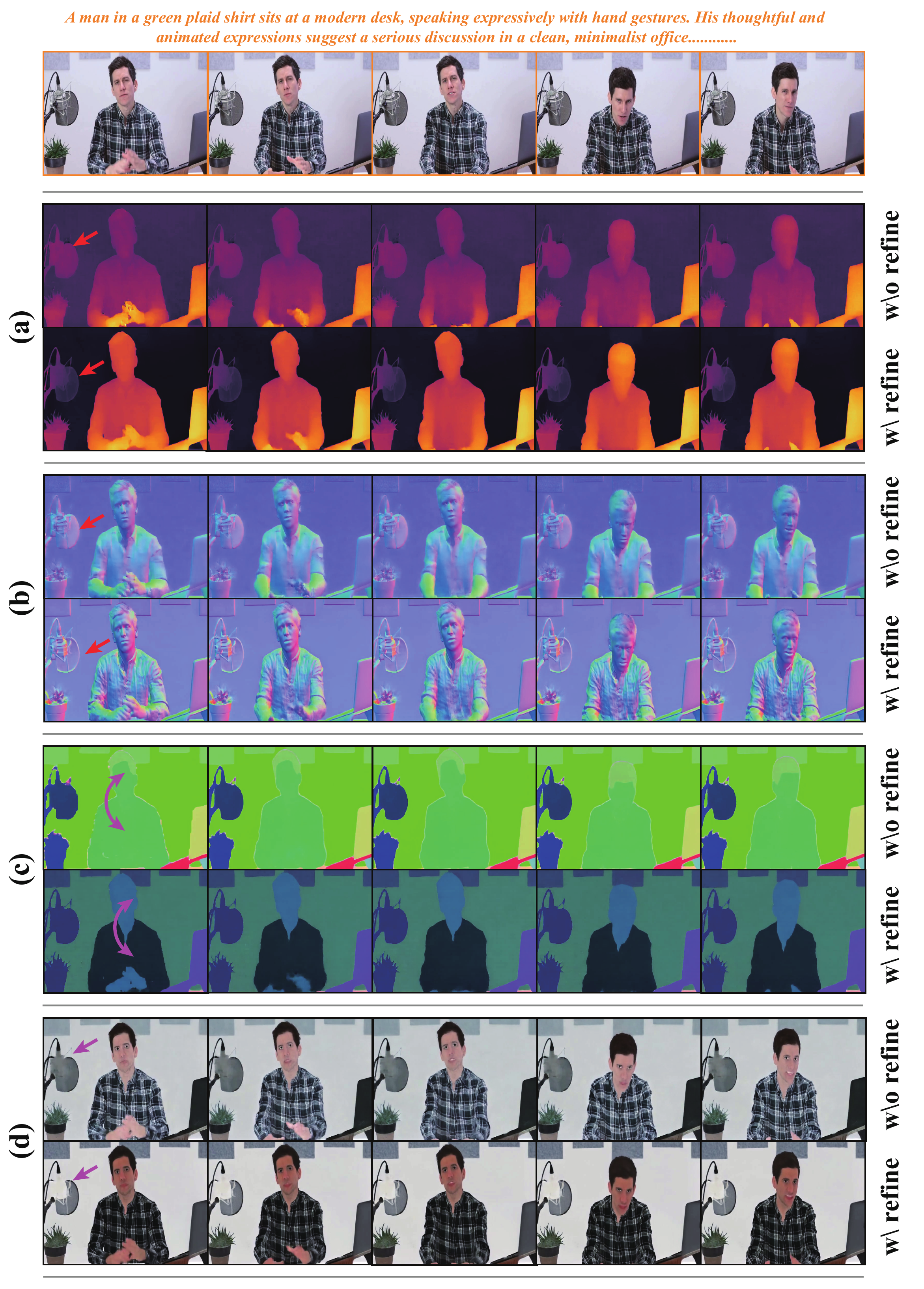}

\caption{
\textbf{Qualitative ablation on the refine stage.} 
With the same input, we evaluate (a) \textit{depth}, (b) \textit{normal}, (c) \textit{segmentation}, and (d) \textit{material (albedo)} with and without refine stage.(\textcolor[HTML]{f7931e}{\boldmath$\rightarrow$} and \textcolor[HTML]{ff1d25}{\boldmath$\rightarrow$} indicate the improvements brought by our refine stage.)}
\vspace{-0.4cm}
  \label{fig:ablation}
\end{figure}

We conduct an ablation study to evaluate the contribution of a key design component—the self-augmentation phase (denoted as \textbf{refine}). 
Both qualitative and quantitative analyses are performed across four modalities: \textit{depth}, \textit{normal}, \textit{segmentation}, and {material (\textit{albedo})}. 
As shown in Table~\ref{tab:ablation} and Figure~\ref{fig:ablation}, incorporating the {refine} stage enables the model to leverage high-quality supervision signals, which improves the overall data quality and enhances both decomposition accuracy and generative capability. The estimations of \textit{depth}, \textit{normal}, and \textit{segmentation} become more accurate, while the \textit{albedo} predictions are more physically plausible after refinement.

\begin{table*}
    \centering
\caption{
\textbf{Ablation study on the effect of the \textit{refine} stage.} 
We evaluate model performance with and without the self-augmentation (\textit{refine}) module across four modalities:
\textit{depth}, \textit{normal}, \textit{segmentation}, and material (\textit{albedo}). 
The \textbf{best} results are highlighted.
}
    \resizebox{0.9\textwidth}{!}{ 
    \begin{tabular}{lccccccccc}
        \toprule
        & \multicolumn{2}{c}{Depth} & \multicolumn{2}{c}{Normal} & \multicolumn{2}{c}{Segmentation} & \multicolumn{3}{c}{Material(Albedo)}  \\
        \cmidrule(lr){2-3} \cmidrule(lr){4-5} \cmidrule(lr){6-7} \cmidrule(lr){8-10}
        Method & AbsRel $\downarrow$ & $\delta_1$ $\uparrow$ & Mean $\downarrow$ & Med $\downarrow$ & M 1-IoU $\uparrow$ & O 1-IoU $\uparrow$  & PSNR $\uparrow$ & SSIM$\uparrow$ & LPIPS $\downarrow$ \\
        \midrule
        w/o refine stage & 0.112 & 0.875 & 15.7 & 7.6 & 52.5 & 68.9 & 22.1 & 0.83 & 0.27 \\
        Full model       & \textbf{0.105} & \textbf{0.889} & \textbf{15.4} & \textbf{7.5} & \textbf{57.0} & \textbf{74.1} & \textbf{23.0} & \textbf{0.88} & \textbf{0.18} \\
        \bottomrule
    \end{tabular}
    }
    \label{tab:ablation}
    \vspace{-0.5cm}
\end{table*}

\subsection{Applications}

As illustrated in Figure~\ref{fig:teaser}(c)--(e), our framework supports a variety of downstream applications, including {prompt-based relighting}, {material editing}, and {object insertion}.

In the relighting scenario, the lighting description in the original text prompt is modified—such as increasing the backlight intensity—while all other modalities remain fixed during generation. 
Benefiting from our multimodal control mechanism, the framework preserves the original scene content and structure throughout the editing process.
  
For material editing (Figure~\ref{fig:teaser}(d)), we modify local \textit{albedo} values, such as regions on the hands, shoes, and text areas (``com''). 
Meanwhile, all other modalities remain unchanged, resulting in precise localized edits, while unedited regions stay consistent with original input video.


Meanwhile, object insertion (Figure~\ref{fig:teaser}(e)) leverages mask regions from the generated {segmentation} modality. 
By editing the {albedo} and {normal} maps and using them as conditions, a bottle and a bowl are seamlessly inserted into the scene. 
These examples demonstrate the flexibility and fine-grained controllability of our unified framework across video understanding and generation tasks.
\vspace{-0.1cm}

\section{Conclusion}

In this work, we introduced \abbname{}, a unified diffusion framework that simultaneously tackles video understanding and controllable video generation. 
By leveraging \hmcs{} and training on the multimodal \mydata{} dataset, our model integrates geometric, appearance, structure, and semantic cues and allows precise and interpretable video control. The framework supports diverse physically meaningful edits and achieves promising results in both understanding and generation.


For future work, we aim to pursue finer-grained control over appearance and geometry, enable explicit light-source manipulation for relighting, and incorporate stronger pre-trained diffusion models such as WAN~\cite{wan2025} to further improve fidelity and controllability.

{
    \small
    \bibliographystyle{ieeenat_fullname}
    \bibliography{main}
}

\clearpage
\setcounter{page}{1}
\maketitlesupplementary

\section{Implementation Details}

\subsection{Train Details}
We base our work on CogVideoX~\cite{yang2024cogvideox}, a T2V diffusion framework, and adopt \textit{CogVideoX1.5-5B} as the foundation for model adaptation. 
Training is performed for a total of \textbf{101K} iterations with a lr of  \(2\times10^{-5}\), using \textbf{8} NVIDIA H100 GPUs and a batch size of \textbf{8}. 
The first and second stages are each trained for \textbf{50K} steps, followed by an additional \textbf{1K} refinement steps in the third stage. 
To achieve efficient large-scale optimization and reduce memory overhead, we employ the {DeepSpeed ZeRO-2} configuration for distributed data-parallel training.

\subsection{Modality Specific Projection Layer.}
To address the heterogeneity among different visual modalities, we introduce a set of modality-specific projection layers, each independently parameterized to project modality-specific features into a shared latent space. As shown in Figure~\ref{fig:method_overview}, each modality is assigned an individual projection head (indicated in different colors), implemented as a lightweight linear layer followed by a normalization module. During training, these projection layers are re-initialized based on the number of modalities, using the \textit{rgb} modality's parameters as initialization, enabling adaptive feature alignment across modalities while maintaining a unified latent representation and capturing diverse visual characteristics.

\subsection{Details of Train Data}

\paragraph{Synthetic Data in {\mydata{}}.}
For the synthetic portion of {\mydata}, we render all videos as follows.
We employ area lights randomly distributed within the scene center region of $[-4, 4]$, with an interval of $0.5$ between adjacent light sources. This setup creates diverse illumination conditions with varying brightness and shadow patterns, while maintaining consistent lighting intensity across frames to ensure temporal coherence. To diversify motion and viewpoint variation, we design four distinct camera trajectory patterns:
(1) an {arc rotation} trajectory from a randomly sampled point A to point B;
(2) a {linear translation} between A and B;
(3) a {zoom-in/zoom-out} motion centered at a sampled point; and
(4) an {object-centric rotation}, where the camera orbits around a randomly chosen object within a [0°, 180°] range.
The camera height is uniformly sampled within [0.5\,m, 2.0\,m], and 49 poses are captured per trajectory at 16\,FPS. 
All synthetic videos are rendered using the {Cycles Path Tracing} engine in Blender~\cite{Blender} with 128 samples per pixel (SPP). 
To further improve the visual fidelity of rendered RGB sequences, we apply Intel {Open Image Denoise (OIDN)}~\cite{OpenImageDenoise} for post-processing. In total, we render {100K} synthetic video clips following the above configurations, each containing 49 frames at 16\,FPS. 
These rendered sequences provide rich geometric and appearance diversity, forming the core of the synthetic subset in {\mydata{}}.

\paragraph{Real Data in {\mydata{}}.}
For the real-world portion, we generate pseudo-labels for each visual modality as follows. 
\textit{Depth} are estimated using {Video Depth Anything}~\cite{video_depth_anything}, ensuring temporally consistent depth across video frames. 
For \textit{segmentation}, we apply {Semantic-SAM}~\cite{li2023semanticsam} to the first frame for instance segmentation and propagate the masks to subsequent frames with {SAM2}~\cite{sam2}, maintaining both semantic and temporal coherence. 
\textit{Canny} are extracted using the OpenCV implementation of the Canny algorithm~\cite{canny1986computational}. 
The remaining appearance-related modalities—\textit{normal}, \textit{albedo}, \textit{roughness}, and \textit{metallic}—are generated via {DiffusionRenderer}~\cite{DiffusionRenderer}, which provides physically consistent intrinsic appearance parameters.
In addition, we extend the {InteriorVerse Image Dataset} by enriching it with \textit{segmentation} and \textit{Canny} modalities, and employ {CogVLM}~\cite{yang2024cogvideox} to produce textual captions for each frame, resulting in an additional 50K single-frame video clips.

\paragraph{Refine Data Selection in Stage 3.}
For the data refinement process in Stage~3, we adopt a two-step filtering strategy over the processed 40K video clips. 
We first employ an aesthetic score to select videos with a higher-quality \textit{rgb} appearance. 
Subsequently, we manually filter the samples to ensure that their \textit{albedo}, \textit{roughness}, and \textit{metallic} modalities exhibit physically plausible properties. 
Through this process, we obtain a refined dataset consisting of 20K high-quality video clips.




\subsection{Details of Application}
\begin{figure*}[t]
  \centering
  \includegraphics[width=1.0\textwidth]{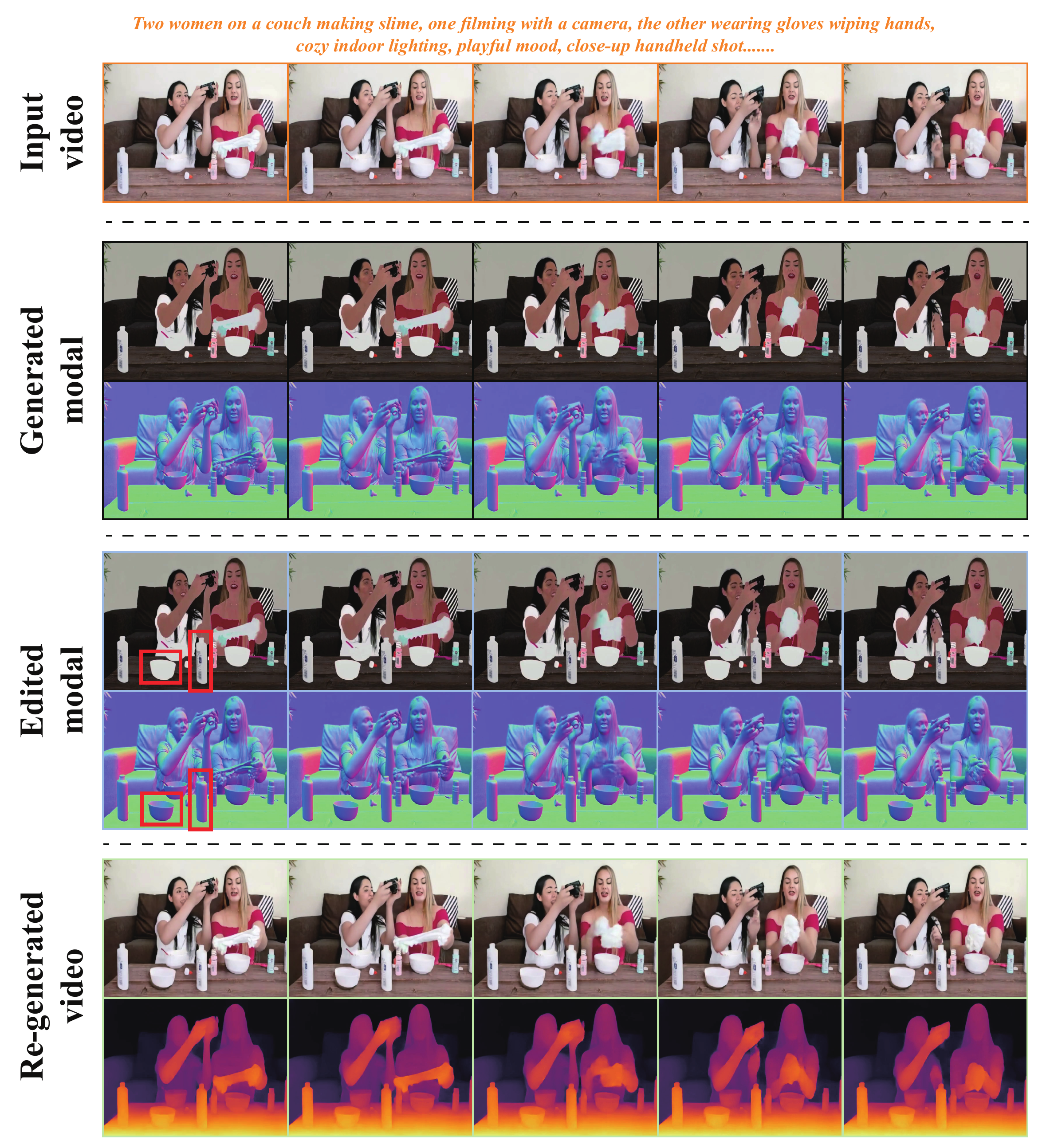}
    \caption{
    \textbf{Implementation details of the object insertion in our application.} We insert objects into the decomposed \textit{albedo} and \textit{normal} modalities and then recompose them to achieve realistic insertion effects, which are particularly evident in the generated \textit{depth} modality.
    }

  \label{fig:supp_application_insert}
\end{figure*}

As illustrated in Figure~\ref{fig:supp_application_insert}, 
our application implements object insertion through a physically grounded reconstruction process. 
We first decompose the input video into intrinsic components, focusing on the \textit{albedo} and \textit{normal} modalities, 
which respectively provide illumination-invariant appearance information and geometric surface structure. 
The target object is then inserted directly into these decomposed modalities, ensuring that its appearance, shading, and geometry remain consistent with the scene’s intrinsic properties. 
After insertion, the modified modalities are recomposed to synthesize the final output. 
This pipeline enables realistic integration of inserted objects into the scene, with the improvement most clearly demonstrated in the generated \textit{depth} modality, 
where structural alignment and geometric coherence are significantly confirmed.


\section{Evaluation Protocol}

\subsection{Depth Evaluation Protocol}
For the depth estimation task, we adopt the evaluation protocol introduced in {Video Depth Anything}~\cite{video_depth_anything}. 
To evaluate the geometric precision of our model, we report the Absolute Relative Error (AbsRel, where lower values indicate better performance) 
and the accuracy threshold $\delta_1$ (where higher values are preferred), consistent with previous works~\cite{hu2024depthcraftergeneratingconsistentlong,yang2024depthv2}. 
Furthermore, to examine the model's zero-shot generalization capability, 
we conduct experiments on the \textbf{ScanNet} dataset~\cite{dai2017scannetrichlyannotated3dreconstructions}, 
which offers diverse real-world indoor scenes for assessing 3D geometric understanding.

\subsection{Segment Evaluation Protocol}
Our segmentation process begins by estimating the first-frame mask using {SemanticSAM}~\cite{li2023semanticsam}, 
which is then propagated across the video sequence using {SAM2}~\cite{sam2} to ensure consistent object tracking. 
For the initial segmentation, we set the granularity level to 2, making the quality of the first-frame mask crucial for the overall segmentation accuracy. 
Given this data generation procedure, our comparison experiments focus primarily on the first frame. 
Following the Semantic-SAM protocol, we adopt the {Single-Granularity Interactive Segmentation} setting throughout all evaluations. 

To process the segmentation outputs, we follow the method proposed in {DICEPTION}~\cite{zhao2025diceptiongeneralistdiffusionmodel}, 
where {K-Means clustering} is applied to generate class-specific masks. 
For each predicted mask, we compute the {Intersection over Union (IoU)} with the corresponding ground-truth mask 
from the {COCO 2017 Val} dataset~\cite{lin2015microsoftcococommonobjects}. 

\subsection{Normal Evaluation Protocol}
Following the evaluation setup of {NormalCrafter}~\cite{bin2025normalcrafterlearningtemporallyconsistent}, 
we perform an extensive analysis of \abbname{} using two well-established benchmarks. 
For the \textit{Sintel} dataset, we leverage the temporally contiguous frames split protocol proposed by {DSINE}~\cite{bae2024rethinkinginductivebiasessurface}. 
On the \textit{ScanNet} dataset, we select 20 unique scenes, each containing 50 sequential frames, 
allowing a balanced assessment of both frame-wise stability and detailed normal prediction accuracy. 
Adhering strictly to the {DSINE} evaluation protocol, 
we compute the angular deviation (measured in degrees) between the predicted surface normals and the corresponding ground truth. 
We report the {mean} and {median} angular errors (lower is better), 
along with the percentage of pixels whose angular errors fall below thresholds of 11.25°, 22.5°, and 30° (higher is better).

\subsection{Material Evaluation Protocol}

Following {DiffusionRenderer}~\cite{DiffusionRenderer} and {RGBX}~\cite{Zeng_2024}, 
we evaluate material estimation using {PSNR}, {SSIM}, and {LPIPS}~\cite{zhang2018unreasonableeffectivenessdeepfeatures}.
We conduct quantitative comparisons with baseline methods on the indoor scene benchmark {InteriorVerse}~\cite{zhu2022learning}.

\subsection{Video Generation Evaluation Protocol}
In alignment with the evaluation framework proposed in {OmniVDiff}~\cite{xdb2025OmniVDiff}, 
we adopt {VBench} as the principal benchmark to systematically assess the quality of our video generation results.  
We evaluate the generated videos along six fundamental dimensions:
\begin{enumerate}
    \item \textbf{Background Consistency:} Evaluates the spatial steadiness and structural coherence of background regions.
    \item \textbf{Dynamic Degree:} Measures the magnitude and richness of dynamic movement throughout the video sequence.
    \item \textbf{Aesthetic Quality:} Examines the visual composition, artistic impression, and overall aesthetic appeal.
    \item \textbf{Motion Smoothness:} Assesses the naturalness, continuity, and perceptual realism of both object and camera motions.
    \item \textbf{Imaging Quality:} Determines the sharpness, clarity, and rendering fidelity of individual video frames.
    \item \textbf{Subject Consistency:} Quantifies the temporal alignment and identity preservation of major subjects across frames.
\end{enumerate}

The final {VBench} score is computed as a weighted combination of these six criteria, 
following the official weighting policy:
\begin{itemize}
    
    \item Background Consistency: 1.0
    \item Dynamic Degree: 0.5
    \item Aesthetic Quality: 1.0
    \item Motion Smoothness: 1.0
    \item Imaging Quality: 1.0
    \item Subject Consistency: 1.0
\end{itemize}

For quantitative assessment, we randomly select 2,048 samples from the validation set to compute each sub-metric independently.

\section{Additional Experiments}

\subsection{Material Estimation}

\begin{figure}[htbp]
    \centering
    \includegraphics[width=1.0\linewidth]{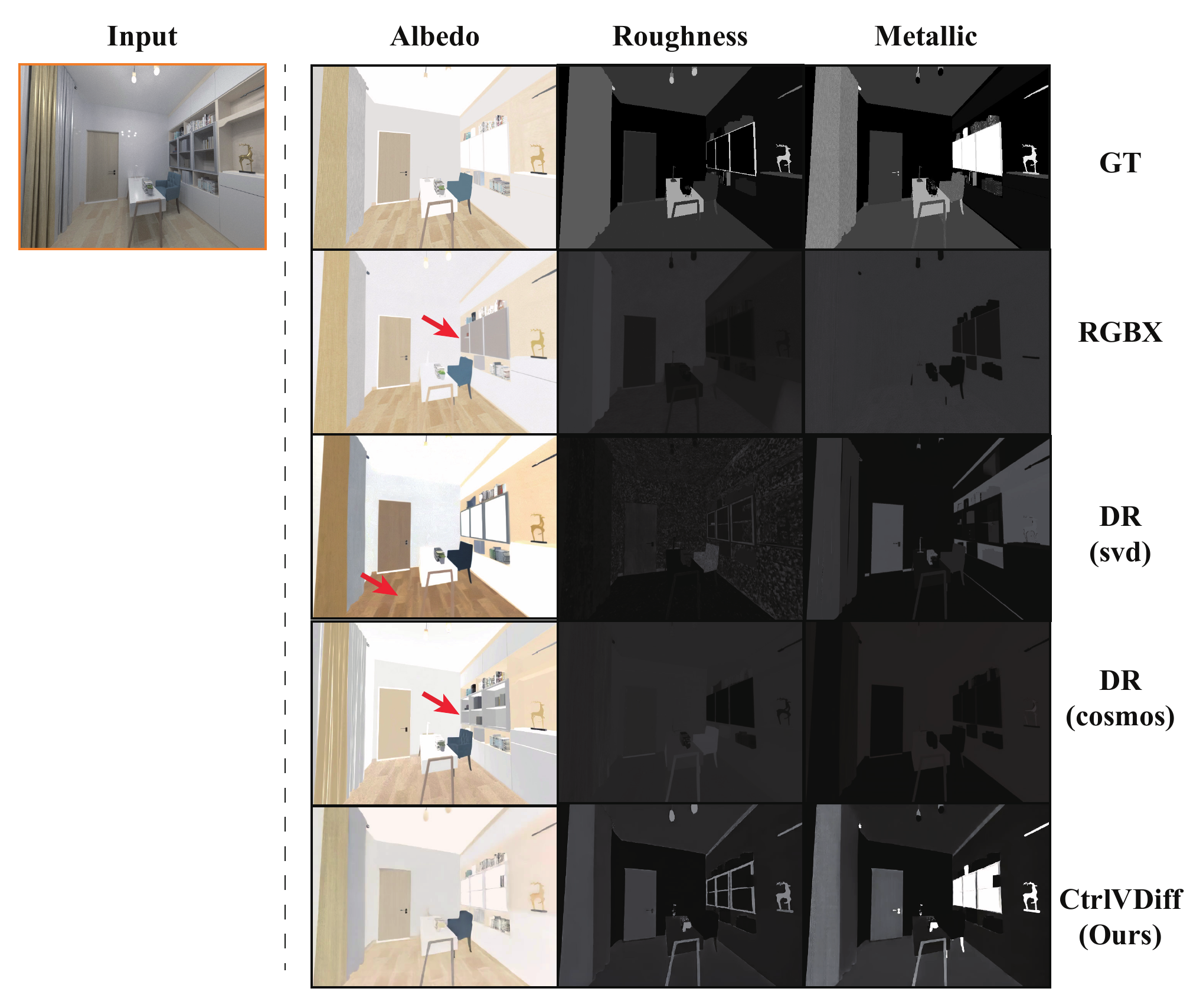}
\caption{
\textbf{Qualitative comparison of material estimation.} 
We evaluate all methods on the {InteriorVerse test dataset}. 
{DiffusionRenderer} is denoted as {DR}. 
All four approaches are designed for indoor scenes. 
Benefiting from our carefully constructed dataset, 
\abbname{} achieves more accurate \textit{albedo} estimation 
(as indicated by the \textcolor[HTML]{ff1d25}{\boldmath$\rightarrow$} in the figure) 
and demonstrates significantly higher accuracy in predicting \textit{roughness} and \textit{metallic} properties compared with existing methods.
}
\label{fig:est_material}
\end{figure}

As shown in Table~\ref{tab:est_material} and Figure~\ref{fig:est_material}, 
{RGBX}~\cite{Zeng_2024} struggles with \textit{albedo} estimation on specular surfaces and exhibits inaccuracies in predicting \textit{roughness} and \textit{metallic} properties. 
{DiffusionRenderer (Cosmos)}~\cite{DiffusionRenderer} provides generally reasonable estimations, 
but the results suffer from noticeable \textit{noise}, particularly in regions such as curtains and floors, 
and display imprecision in the \textit{roughness} and \textit{metallic} channels. 
{DiffusionRenderer (SVD)}~\cite{DiffusionRenderer} produces smoother results but deviates from ground truth, 
introducing substantial noise in the \textit{roughness} and \textit{metallic} maps, 
which degrades overall accuracy.

Although {DiffusionRenderer (Cosmos)} serves as our expert model, 
the Stage~3 training strategy effectively mitigates the noise and bias introduced by pseudo-labels, 
enabling \abbname{} to achieve the best estimation performance across all three material channels. 
In particular, our method shows a significant improvement in predicting \textit{roughness} and \textit{metallic} properties, 
outperforming all existing baselines by a clear margin.

\begin{table*}
    \centering
    \label{tab:est_material}
    \resizebox{\textwidth}{!}{ 
    \begin{tabular}{lccccccccc}
        \toprule
        & \multicolumn{3}{c}{Albedo} & \multicolumn{3}{c}{Metallic} & \multicolumn{3}{c}{Roughness} \\
        \cmidrule(lr){2-4} \cmidrule(lr){5-7} \cmidrule(lr){8-10}
        Method & PSNR $\uparrow$ & SSIM$\uparrow$ & LPIPS $\downarrow$ & PSNR $\uparrow$ & SSIM$\uparrow$ & LPIPS $\downarrow$  & PSNR $\uparrow$ & SSIM$\uparrow$ & LPIPS $\downarrow$ \\
        \midrule
        
        RGB+X~\cite{Zeng_2024} & 16.4 & 0.78 & \underline{0.19}  &12.8 & 0.51 & 0.54 & 10.6 & 0.46 & 0.64  \\
        
        DR(cosmos)~\cite{DiffusionRenderer}  & 18.4 & 0.80 & 0.32 & 12.2 & 0.46 & 0.50 & 11.9 & 0.47 & 0.60 \\
        
        DR(svd)~\cite{DiffusionRenderer}     & \underline{22.4} & \underline{0.87} & \underline{0.19} & \underline{13.5} & \underline{0.55} & \underline{0.37} & \underline{14.8} & \underline{0.51} & \underline{0.52} \\
        
        \abbname(Ours)                       & \textbf{23.0} & \textbf{0.88} & \textbf{0.18} & \textbf{14.8} & \textbf{0.62} & \textbf{0.31} & \textbf{15.9} & \textbf{0.65} & \textbf{0.31} \\
        
        \bottomrule
    \end{tabular}
    }
    \caption{
    \textbf{Quantitative Results for Material Property Estimation.} 
    We evaluate the prediction accuracy of \textit{albedo}, \textit{metallic}, and \textit{roughness} on the \textbf{InteriorVerse} benchmark, 
    reporting PSNR, SSIM, and LPIPS scores as evaluation metrics. 
    Our approach consistently outperforms prior works, yielding notable gains in estimating the \textit{roughness} and \textit{metallic} components 
    compared with RGBX~\cite{Zeng_2024} and DiffusionRenderer~\cite{DiffusionRenderer}. 
    The \textbf{top-performing} and \underline{second-best} results are indicated for clarity.
    }
    \label{tab:est_material}
\end{table*}

\subsection{Analysis of Video Generation under Selected Condition Combinations}

Our framework supports a wide range of condition combinations for video generation. 
To better analyze the performance differences among multiple combinations, 
we explicitly group the decomposed modalities into four categories based on their inherent characteristics: 
\textit{geometry} (\textit{depth} + \textit{normal}), 
\textit{appearance} (\textit{albedo} + \textit{roughness} + \textit{metallic}), 
\textit{semantic} (\textit{segmentation}), and 
\textit{structure} (\textit{Canny}). 
Since \textit{segmentation} captures the categorical composition of objects in the scene, 
and \textit{Canny} primarily encodes the structural layout, 
these two modalities together describe the overall scene configuration. 
Therefore, we further merge them into a unified \textit{layout} category. 

Consequently, the eight modalities are grouped into three major sets: 
\textit{geometry}, \textit{appearance}, and \textit{layout}. 
We systematically combine these sets and conduct both qualitative and quantitative analyses 
to assess their respective impacts on video generation quality.

\begin{figure*}[t]
  \centering
  \includegraphics[width=1.0\textwidth]{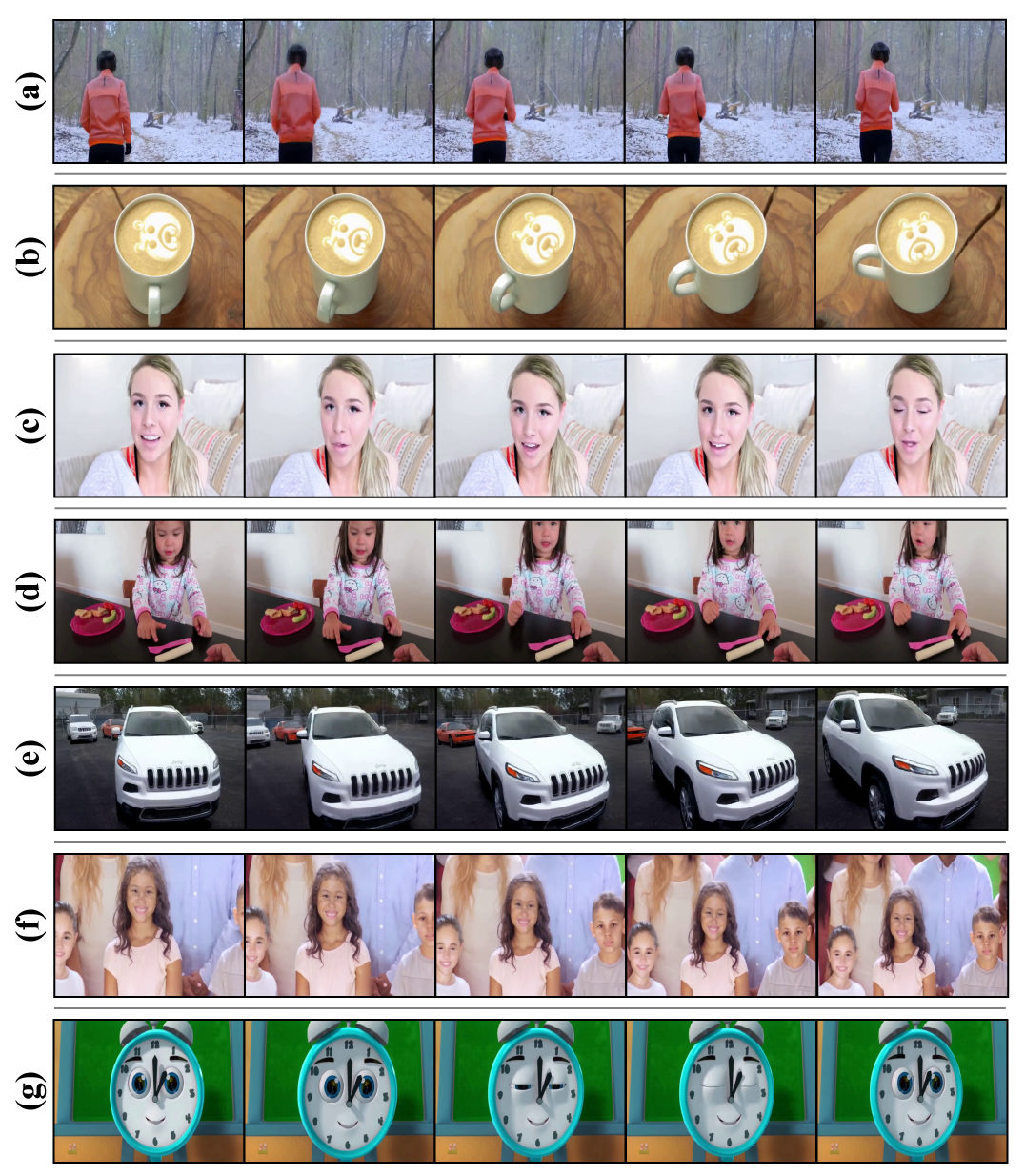}
\caption{
\textbf{Qualitative Analysis of Video Generation under Selected Condition Combinations.}
Visualization of \abbname{} generating videos conditioned on various modality combinations: 
(a) geometry, 
(b) appearance, 
(c) layout, 
(d) geometry + appearance, 
(e) geometry + layout, 
(f) layout + appearance, and 
(g) geometry + appearance + layout. 
The results demonstrate that \abbname{} effectively adapts to different condition combinations for diverse video generation scenarios.
}
\label{fig:videogen_analysis_cond}
\end{figure*}

As presented in Table~\ref{tab:videogen_analysis_cond} and Figure~\ref{fig:videogen_analysis_cond}, 
all combinations yield visually coherent and high-quality results, 
demonstrating the strong adaptability of our model to various condition configurations. 
Overall, we observe that combinations involving \textit{appearance} modalities tend to produce superior visual realism, 
as reflected in Figure~\ref{fig:videogen_analysis_cond} (g), (f), (b), and (d). 
In contrast, as shown in Table~\ref{tab:videogen_analysis_cond} and Figure~\ref{fig:videogen_analysis_cond} (e) and (f), 
when compared with (a) and (b), the inclusion of \textit{layout}-related modalities 
(as in cases (e) and (f)) yields only modest improvements, 
indicating that spatial guidance alone provides limited enhancement to perceptual quality.
\begin{table*}[t]
\centering
\resizebox{\textwidth}{!}{
\begin{tabular}{lccccccc}
\toprule
Model & subject consistency & b.g. consistency & motion smoothness & dynamic degree & aesthetic quality & imaging quality & weighted average \\

\midrule
\multicolumn{8}{l}{\textit{text+geometry}} \\
\hspace{1em}\abbname(Ours) &\underline{97.83} &{97.14} &{98.92} &{51.85} &{53.60} &{68.11} &{73.59}  \\

\midrule
\multicolumn{8}{l}{\textit{text+appearance}} \\
\hspace{1em}\abbname(Ours) &{97.46} &{97.28} &{99.21} &{50.69} &{57.61} &{70.08} &{74.50}  \\

\midrule
\multicolumn{8}{l}{\textit{text+layout}} \\
\hspace{1em}\abbname(Ours) &{97.52} &{95.74} &{99.17} &{52.75} &{53.17} &{66.96} &{73.16}  \\

\midrule
\multicolumn{8}{l}{\textit{text+geometry+appearance}}  \\
\hspace{1em}\abbname(Ours) &{97.64} &\underline{97.49} &\underline{99.41} &{53.48} &\underline{57.82} &{70.87} &{75.00}  \\

\midrule
\multicolumn{8}{l}{\textit{text+geometry+layout}} \\
\hspace{1em}\abbname(Ours) &{97.31} &{96.66} &{98.77} &\underline{53.91} &{53.89} &{68.29} &{73.65}  \\

\midrule
\multicolumn{8}{l}{\textit{text+layout+apparance}} \\
\hspace{1em}\abbname(Ours) &{97.62} &{97.28} &{99.08} &{53.39} &{57.62} &\underline{70.55} &{74.81} \\

\midrule
\multicolumn{8}{l}{\textit{text+all}} \\
\hspace{1em}\abbname(Ours) &\textbf{98.73} &\textbf{98.46} &\textbf{99.49} &\textbf{54.25} &\textbf{58.72} &\textbf{71.63} &{75.69}  \\
\bottomrule
\end{tabular}
}
\caption{
\textbf{Evaluation of Video Generation Across Selected Condition Combinations.} 
For clarity, the \textbf{bold} values represent the best performance, while the \underline{underlined} ones indicate the second-best results.
}
\label{tab:videogen_analysis_cond}
\end{table*}

\subsection{Single Condition Video Generation}
\begin{figure*}[t]
  \centering
  \includegraphics[width=1.0\textwidth]{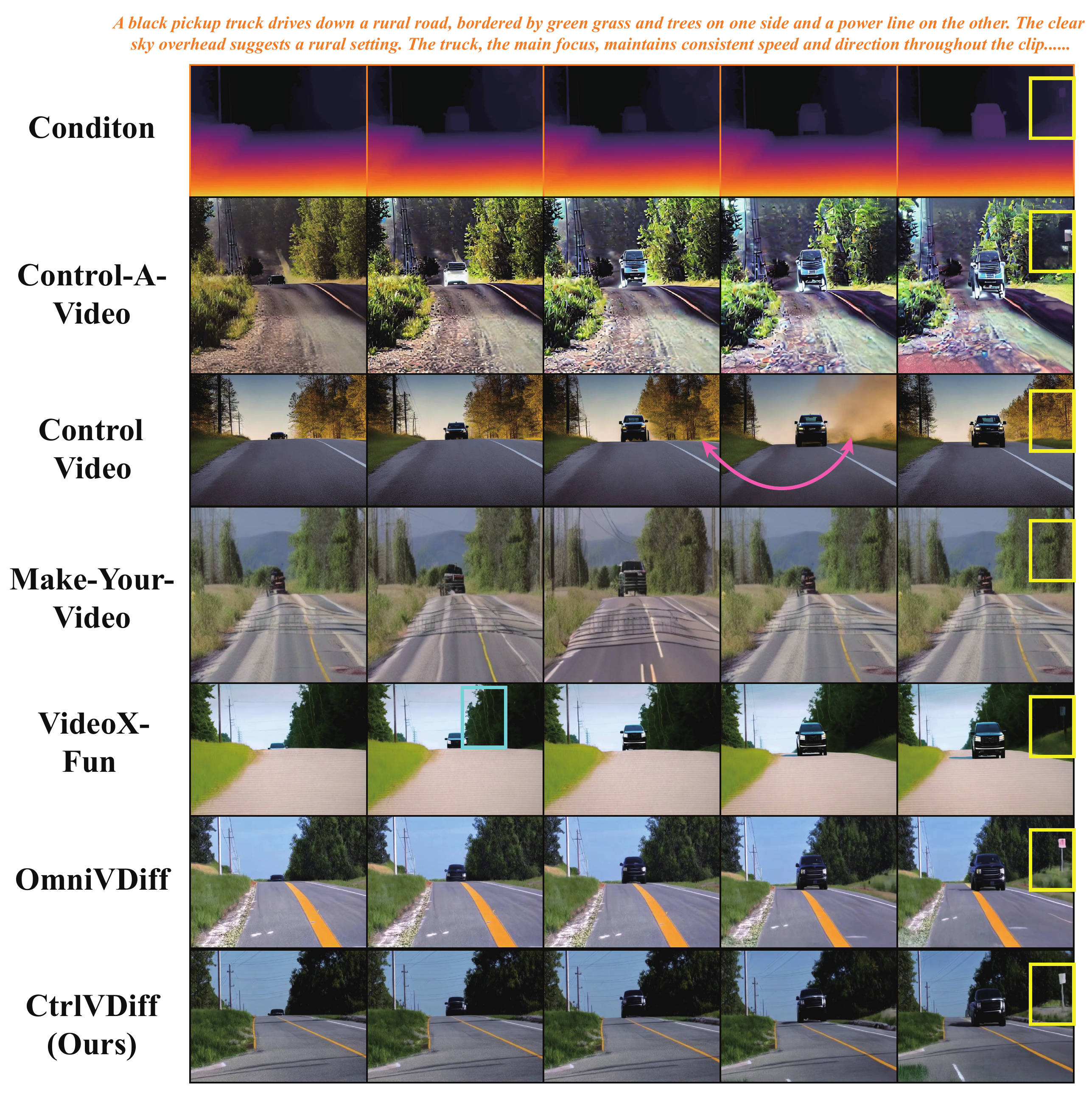}
\caption{
\textbf{Qualitative results of depth-conditioned video synthesis.} 
Regions outlined in \textcolor{yellow}{yellow} indicate areas where our approach preserves depth consistency more effectively than competing methods. 
The \textcolor{pink}{pink arrows} denote temporal discontinuities across frames, 
and the \textcolor{cyan}{cyan boxes} point to visible distortions in the RGB sequences. 
Overall, our method exhibits enhanced temporal coherence and improved visual fidelity.
}

  \label{fig:single_cond_video_gen_depth}
\end{figure*}

\begin{figure*}[t]
  \centering
  \includegraphics[width=1.0\textwidth]{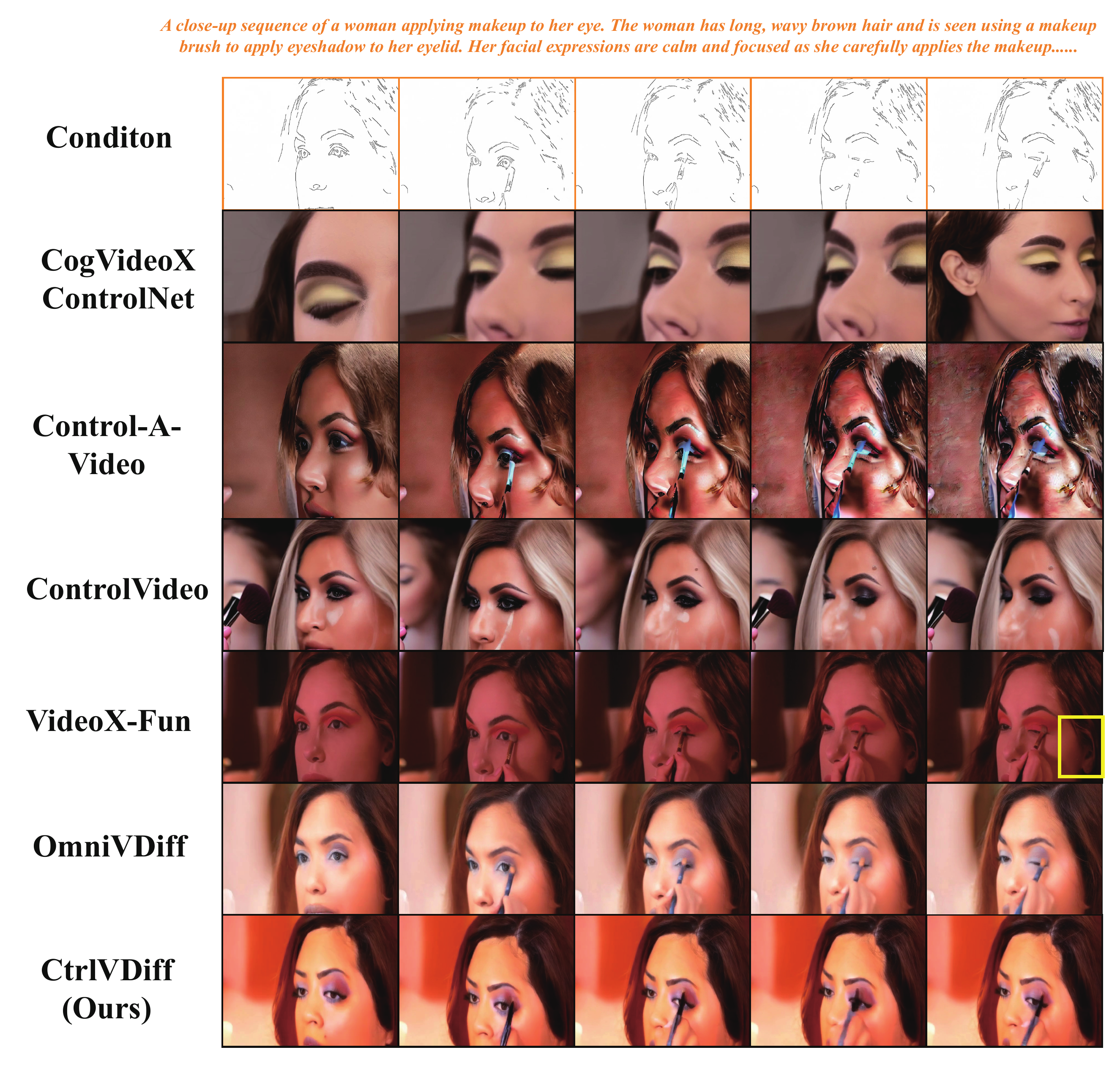}
\caption{
\textbf{Qualitative results of Canny-conditioned video generation.}
Regions highlighted with \textcolor{yellow}{yellow boxes} reveal visual artifacts in the baseline RGB outputs. 
Unlike these baselines, our model produces results that better conform to the provided \textit{Canny} edge conditions, 
achieving improved structural precision and enhanced overall visual realism.
}
  \label{fig:single_cond_video_gen_canny}
\end{figure*}

We evaluate our framework in the \textit{single-condition video generation} setting and compare it against task-specific baselines that leverage visual priors such as \textit{depth} and \textit{canny}. 
As presented in Table~\ref{tab:single_cond_video_gen}, Figure~\ref{fig:single_cond_video_gen_depth}, and Figure~\ref{fig:single_cond_video_gen_canny}, 
our method delivers strong performance even under single-modality conditioning, demonstrating clear advantages in maintaining structural integrity and temporal smoothness. 
The quantitative results in Table~\ref{tab:single_cond_video_gen} further show that our framework performs comparably to or surpasses existing approaches under both \textit{depth}- and \textit{canny}-guided settings. 
Empowered by the unified diffusion backbone, \abbname{} enables controllable and flexible video generation across diverse modalities within a single generative system. 

As reported in Table~\ref{tab:single_cond_video_gen}, the performance under the \textit{metallic} condition appears relatively weaker compared to other modalities. 
We attribute this to the inherently sparse spatial distribution of \textit{metallic} information, which provides less effective conditional guidance during the generative process.

In contrast, \textit{normal}-, \textit{roughness}-, and \textit{albedo}-conditioned generations exhibit highly competitive results, 
with the \textit{albedo}-guided setup consistently outperforming all baselines across most quantitative and qualitative metrics. 
This superiority can be primarily attributed to the strong appearance-controlling capability of the \textit{albedo} modality, 
which enhances overall visual quality, particularly in terms of \textit{aesthetic quality} and \textit{imaging quality}.

\begin{table*}[!htbp]
\centering
\resizebox{\textwidth}{!}{
\begin{tabular}{lccccccc}
\toprule
Model & subject consistency & b.g. consistency & motion smoothness & dynamic degree & aesthetic quality & imaging quality & weighted average \\

\midrule
\multicolumn{8}{l}{\textit{text+depth}} \\
\hspace{1em}Control-A-Video\cite{chen2023control} & 89.99 & 91.63 & 91.90 & 40.62 & 48.67 & 68.69 & 68.53 \\
\hspace{1em}ControlVideo\cite{zhang2023controlvideo} & 95.50 & 94.17 & 97.80 & 18.35 & \textbf{57.56} & \underline{70.09} & 70.71 \\
\hspace{1em}Make-your-video\cite{xing2024make} & 90.04 & 92.48 & 97.64 & {51.95} & 44.67 & \textbf{70.26} & 70.17 \\
\hspace{1em}VideoX-Fun\cite{VideoXFun2024} & {96.25} & \underline{95.73} & \underline{98.90} & 50.43 & \underline{55.81} & 55.38 & {72.85} \\

\hspace{1em}\before~\cite{xdb2025OmniVDiff} & \textbf{97.96} & \textbf{96.66} & \underline{99.18} & \textbf{53.32} & 52.95 & 67.26 & \textbf{73.45} \\

\hspace{1em}\abbname(Ours)  &\underline{97.56} &\underline{96.36} &\textbf{99.30} & \underline{52.80} &{52.73} &{67.56} &\underline{73.32} \\

\midrule
\multicolumn{8}{l}{\textit{text+canny}} \\
\hspace{1em}CogVideoX+CTRL\cite{cogvideoxControlNet2024} & 96.26 & 94.53 & 98.42 & \underline{53.44} & 49.34 & 55.56 & 70.13 \\
\hspace{1em}Control-A-Video\cite{chen2023control} & 89.81 & 91.27 & 97.86 & 41.79 & 47.23 & \textbf{68.77} & 69.31 \\
\hspace{1em}ControlVideo\cite{zhang2023controlvideo} & 95.23 & 94.00 & 97.12 & 17.58 & \textbf{55.81} & 55.38 & 67.72 \\
\hspace{1em}VideoX-Fun\cite{VideoXFun2024} & \underline{96.69} & \underline{95.41} & \underline{99.15} & 50.78 & \underline{52.99} & 66.76 & \underline{72.73} \\
\hspace{1em}\before~\cite{xdb2025OmniVDiff} & \textbf{97.84} & \textbf{95.55} & \textbf{99.23} & \textbf{53.53} & 52.34 & \underline{67.14} & \textbf{73.14} \\
\hspace{1em}\abbname(Ours) &{97.61} &{95.37} &{99.11} &{53.23} &{51.89} &{67.20} &{72.97} \\

\midrule
\multicolumn{8}{l}{\textit{text+segment}} \\
\hspace{1em}\before~\cite{xdb2025OmniVDiff} & \textbf{97.97} & \textbf{95.81} & \textbf{99.31} & \textbf{53.18} & \textbf{53.37} & \textbf{67.51} & \textbf{73.42} \\
\hspace{1em}\abbname(Ours) &\underline{97.52} &\underline{95.74} &\underline{99.21} &\underline{52.75} &\underline{53.17} &\underline{66.96} &\underline{73.16} \\

\midrule
\multicolumn{8}{l}{\textit{text+normal}} \\
\hspace{1em}\abbname(Ours) &\textbf{97.72} &\textbf{95.87} &\textbf{99.19} &\textbf{52.19} &\textbf{51.9} &\textbf{67.45} &\textbf{73.04} \\

\midrule
\multicolumn{8}{l}{\textit{text+albedo}} \\
\hspace{1em}\abbname(Ours) &\textbf{97.34} &\textbf{97.17} &\textbf{98.86} &\textbf{50.90} &\textbf{56.69} &\textbf{69.50} &\textbf{74.17} \\

\midrule
\multicolumn{8}{l}{\textit{text+rogheness}} \\
\hspace{1em}\abbname(Ours) &\textbf{97.79} &\textbf{97.52} &\textbf{97.53} &\textbf{50.37} &\textbf{50.99} &\textbf{67.48} &\textbf{72.75} \\

\midrule
\multicolumn{8}{l}{\textit{text+metallic}} \\
\hspace{1em}\abbname(Ours) &\textbf{96.58} &\textbf{94.91} &\textbf{97.01} &\textbf{48.82} &\textbf{50.18} &\textbf{65.19} &\textbf{71.38} \\


\bottomrule
\end{tabular}

}

\caption{\textbf{VBench metrics for single conditioned video generation.} For each condition type, the best performance is shown in \textbf{bold}, and the second-best is marked with an \underline{underline}.}
\label{tab:single_cond_video_gen}
\end{table*}


\section{More Video Generation Results}
\subsection{Text to Multimodal Video Generation}
\begin{figure*}[t]
  \centering
  \includegraphics[width=1.0\textwidth]{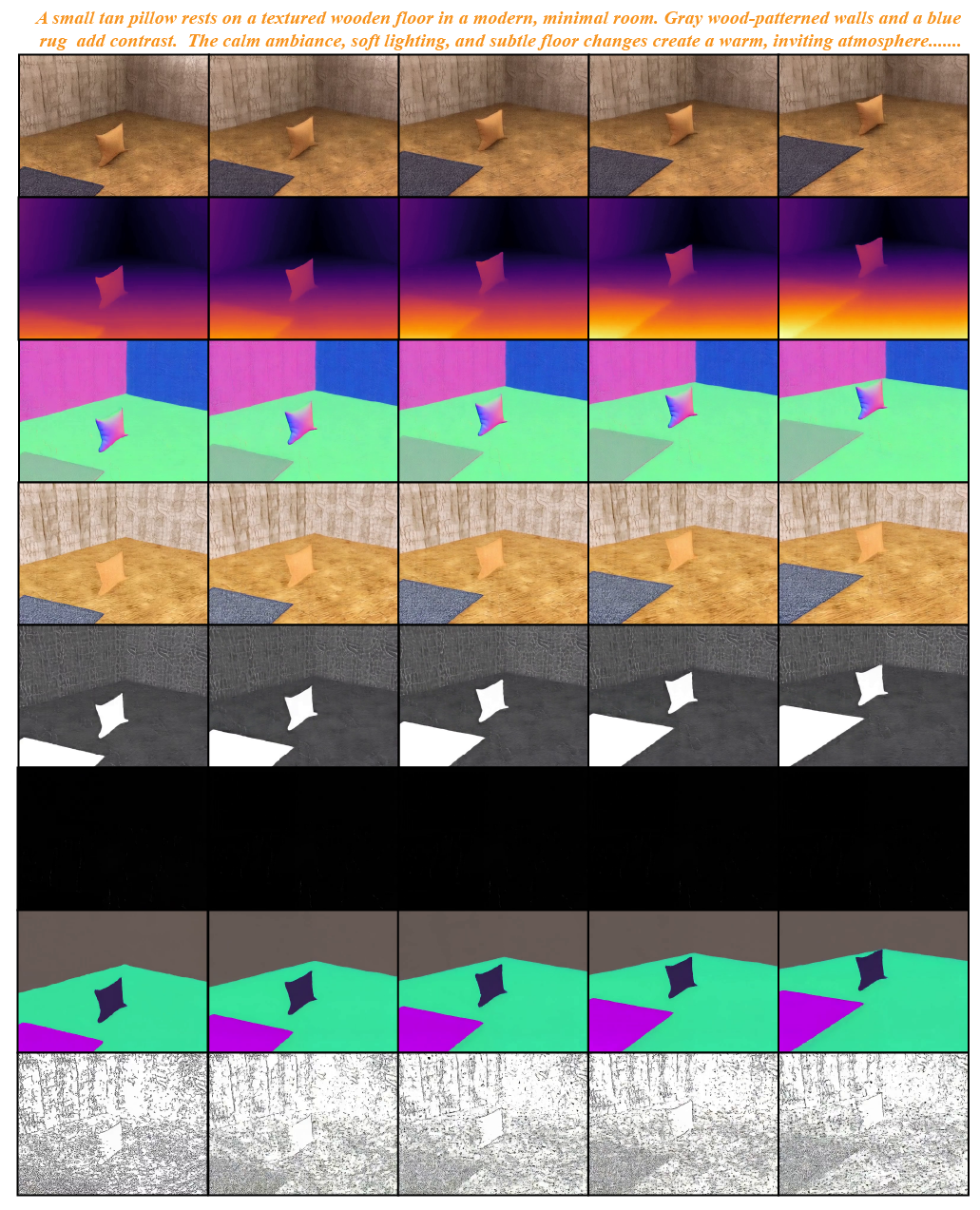}
    \caption{
    \textbf{Additional examples of text-to-multi-modality video generation.}
    Visualization of text-conditioned \textit{synchronous} multi-modality video outputs generated by \abbname{}, 
    demonstrating its ability to produce coherent and semantically aligned visual modalities from textual prompts.
    }
  \label{fig:supp_cond_text}
\end{figure*}

Figure~\ref{fig:supp_cond_text} shows the case where our method generates all modalities conditioned solely on text. The results demonstrate that our approach can produce high-quality rgb videos while simultaneously ensuring the correctness and plausibility of other modalities.

\subsection{Single Condition to Multimodal Video Generation}
\begin{figure*}[t]
  \centering
  \includegraphics[width=1.0\textwidth]{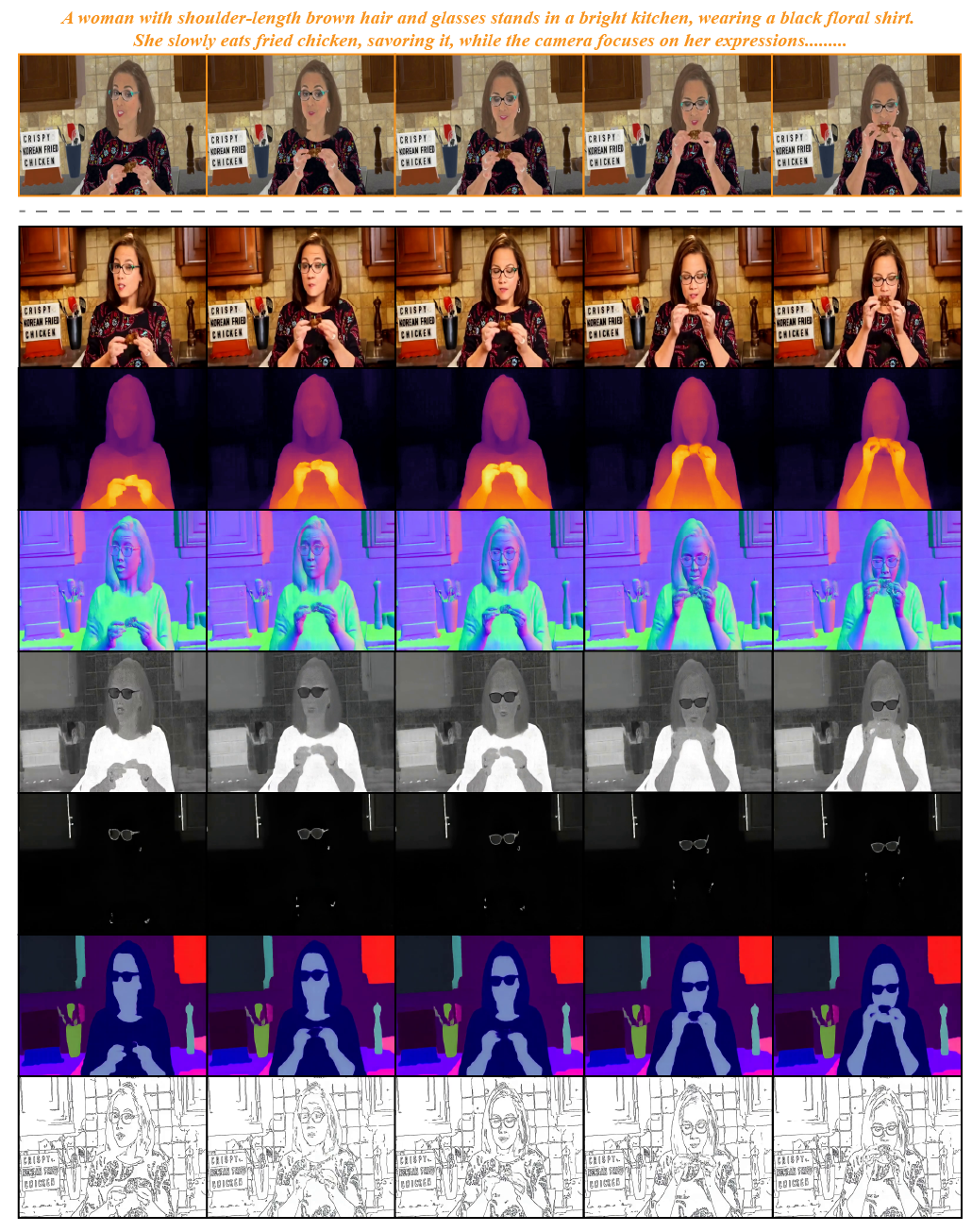}
\caption{
\textbf{Additional qualitative results for single-condition (\textit{albedo}) video generation.}
Visual examples of \abbname{} generating videos conditioned solely on the \textit{albedo} modality,
showcasing precise color reproduction and faithful texture consistency driven by surface reflectance cues.
}
  \label{fig:supp_cond_albedo}
\end{figure*}

\begin{figure*}[t]
  \centering
  \includegraphics[width=1.0\textwidth]{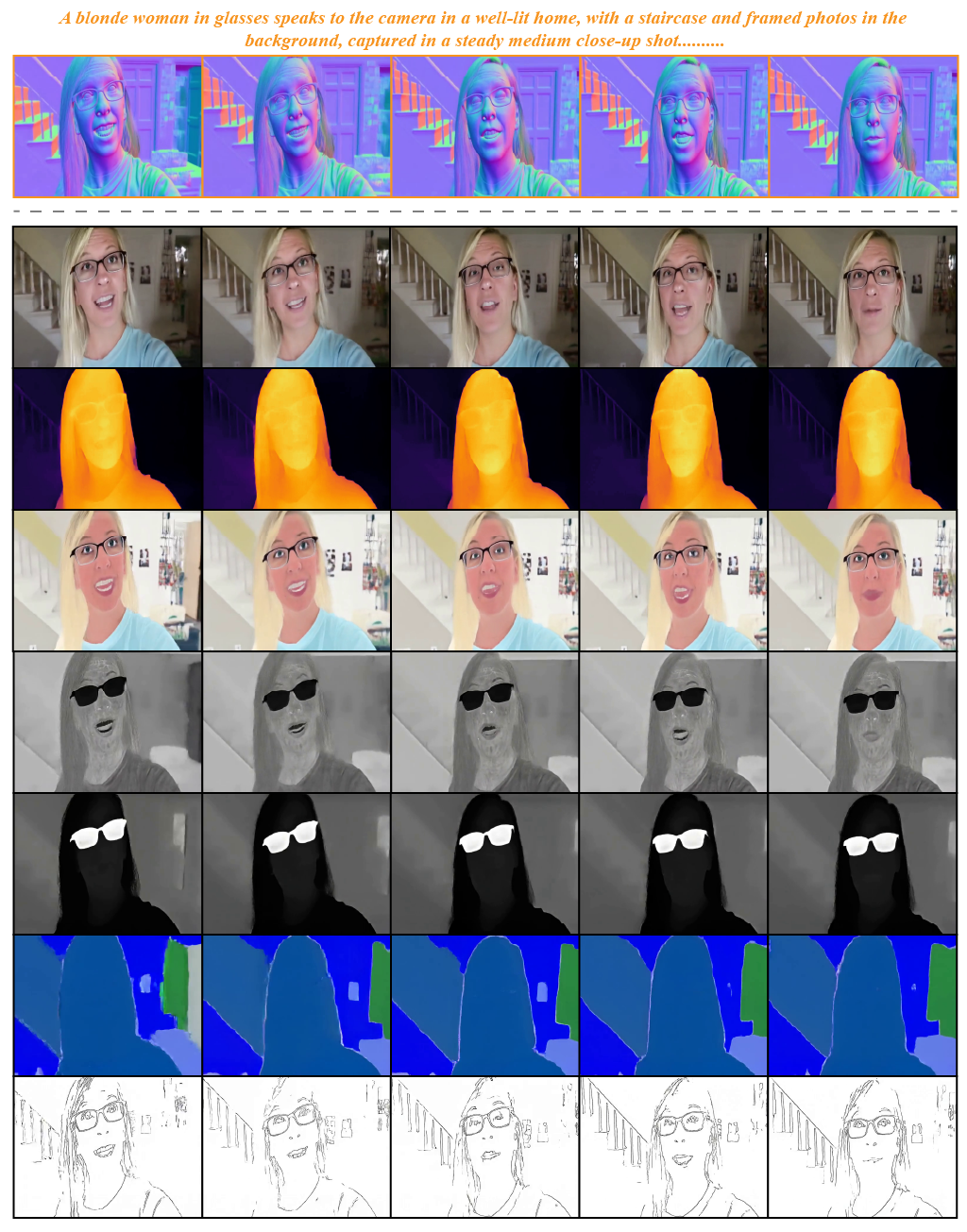}
    \caption{
    \textbf{Additional qualitative examples for single-condition (\textit{normal}) video generation.}
    Visualizations of \abbname{} producing videos conditioned exclusively on the \textit{normal} modality,
    illustrating accurate geometric reconstruction and consistent shading behavior across frames.
    }
  \label{fig:supp_cond_normal}
\end{figure*}

We select \textit{albedo} and \textit{normal} as conditioning inputs for single-condition video generation. Figure~\ref{fig:supp_cond_albedo} shows the results of multimodal generation conditioned on \textit{albedo}, while Figure~\ref{fig:supp_cond_normal} presents the corresponding results using \textit{normal}. These results demonstrate that our method produces stable and consistent outputs under both conditions.

\subsection{Multi Conditions to Multimodal Video Generation}
\begin{figure*}[t]
  \centering
  \includegraphics[width=1.0\textwidth]{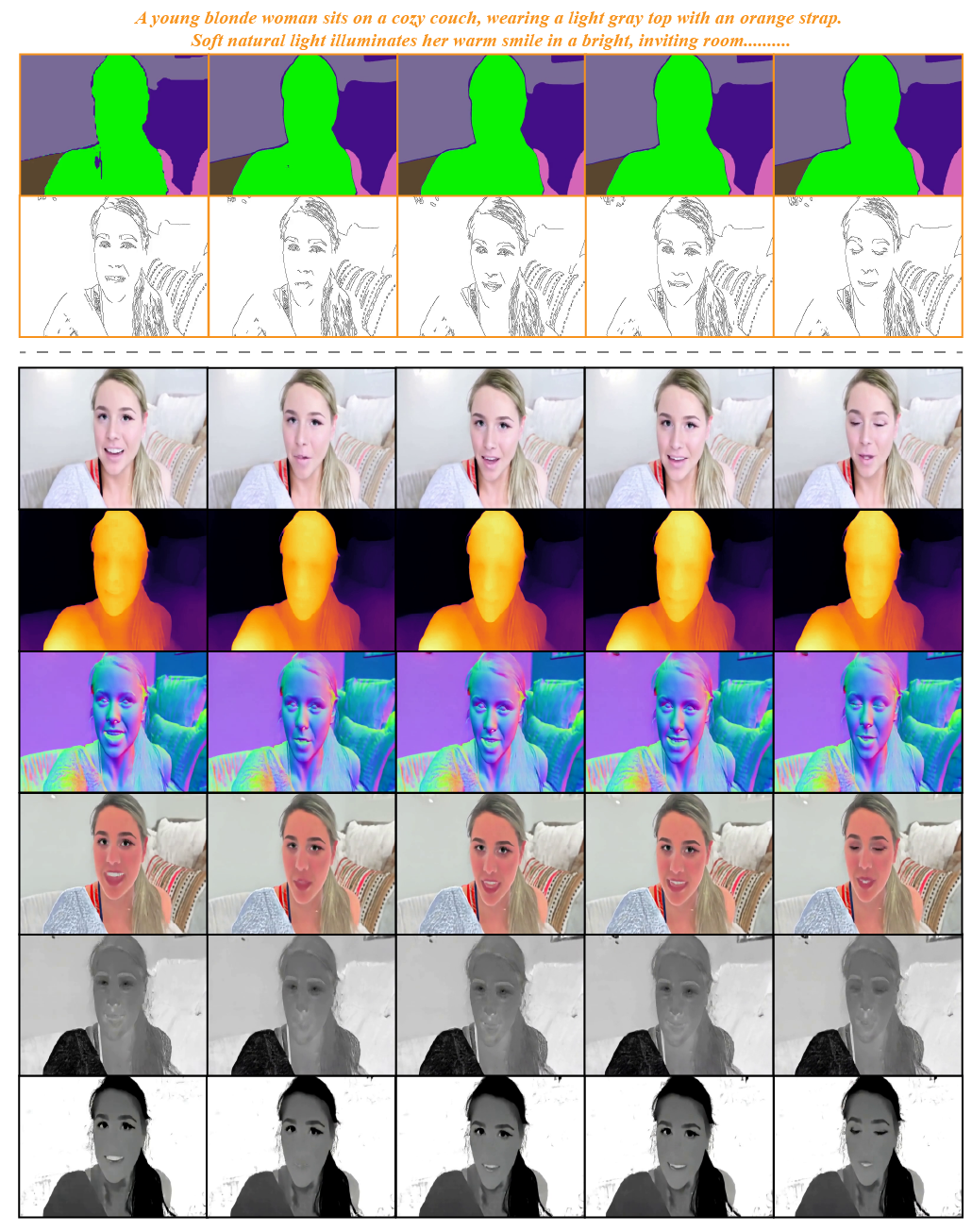}
\caption{
\textbf{Additional qualitative examples of multi-condition (\textit{segmentation} + \textit{canny}) to multi-modality video generation.}
Visualizations of \abbname{} generating \textit{synchronous} multi-modality videos under combined \textit{segmentation} and \textit{canny} conditions. 
Incorporating the \textit{canny} modality alongside \textit{segmentation} provides finer structural constraints, 
resulting in more precise facial geometry and improved temporal coherence across frames.
}
  \label{fig:supp_cond_seg+canny}
\end{figure*}

We present multimodal video generation results conditioned on the combination of \textit{segmentation} and \textit{canny} edges. As shown in Figure~\ref{fig:supp_cond_seg+canny}, the generated videos faithfully adhere to these conditioning signals: the segmentation map guides the layout of semantic regions such as the person, walls, and pillows, while the \textit{canny} edges effectively control the structural details of facial contours and pillow shapes.

\subsection{All Conditions to Multimodal Video Generation}
\begin{figure*}[t]
  \centering
  \includegraphics[width=1.0\textwidth]{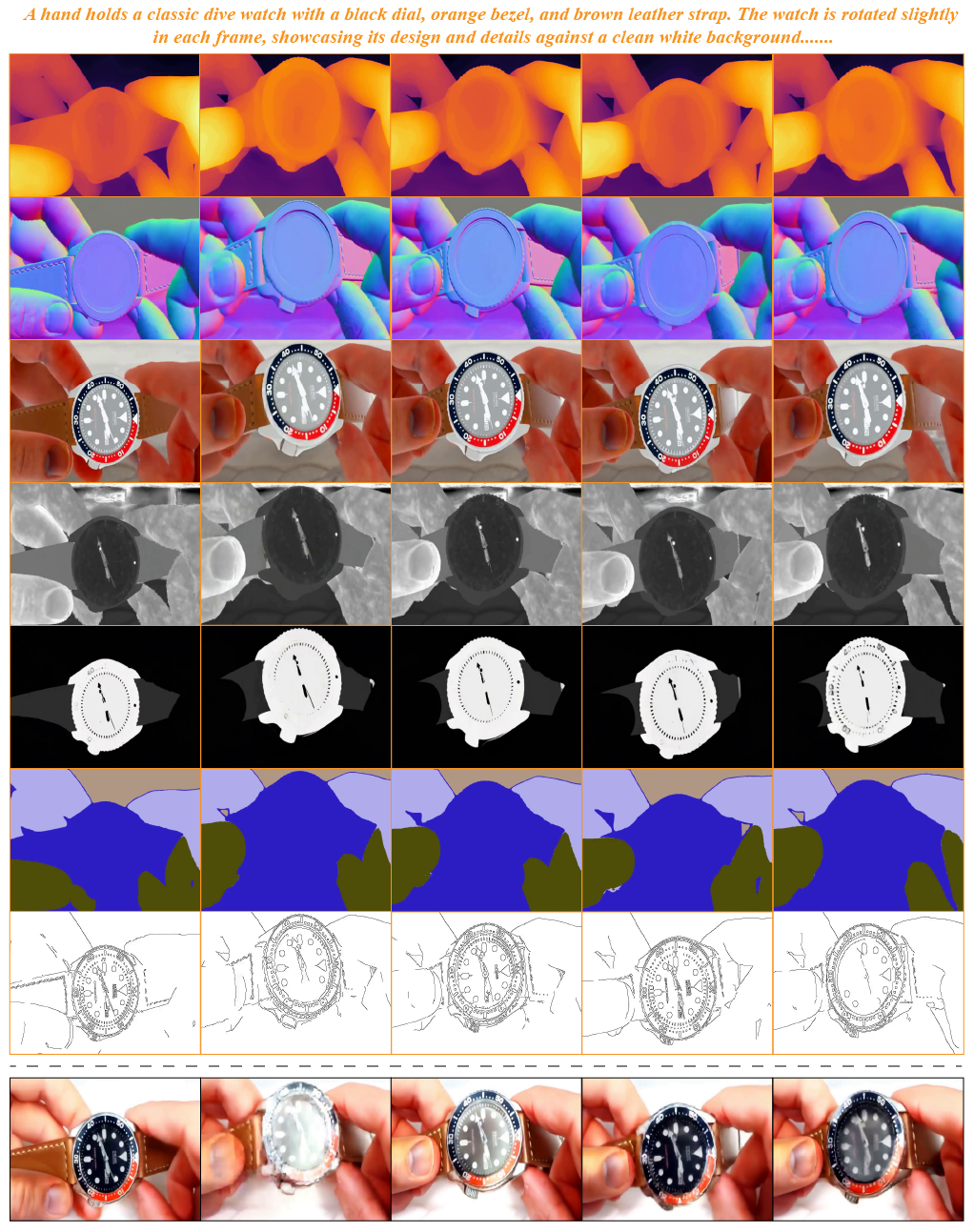}
\caption{
\textbf{Additional qualitative results of video generation.}
Visualizations of \abbname{} demonstrating its multimodal video generation capability across all modality prediction tasks.
}
  \label{fig:supp_cond_all}
\end{figure*}

\begin{figure*}[t]
  \centering
  \includegraphics[width=1.0\textwidth]{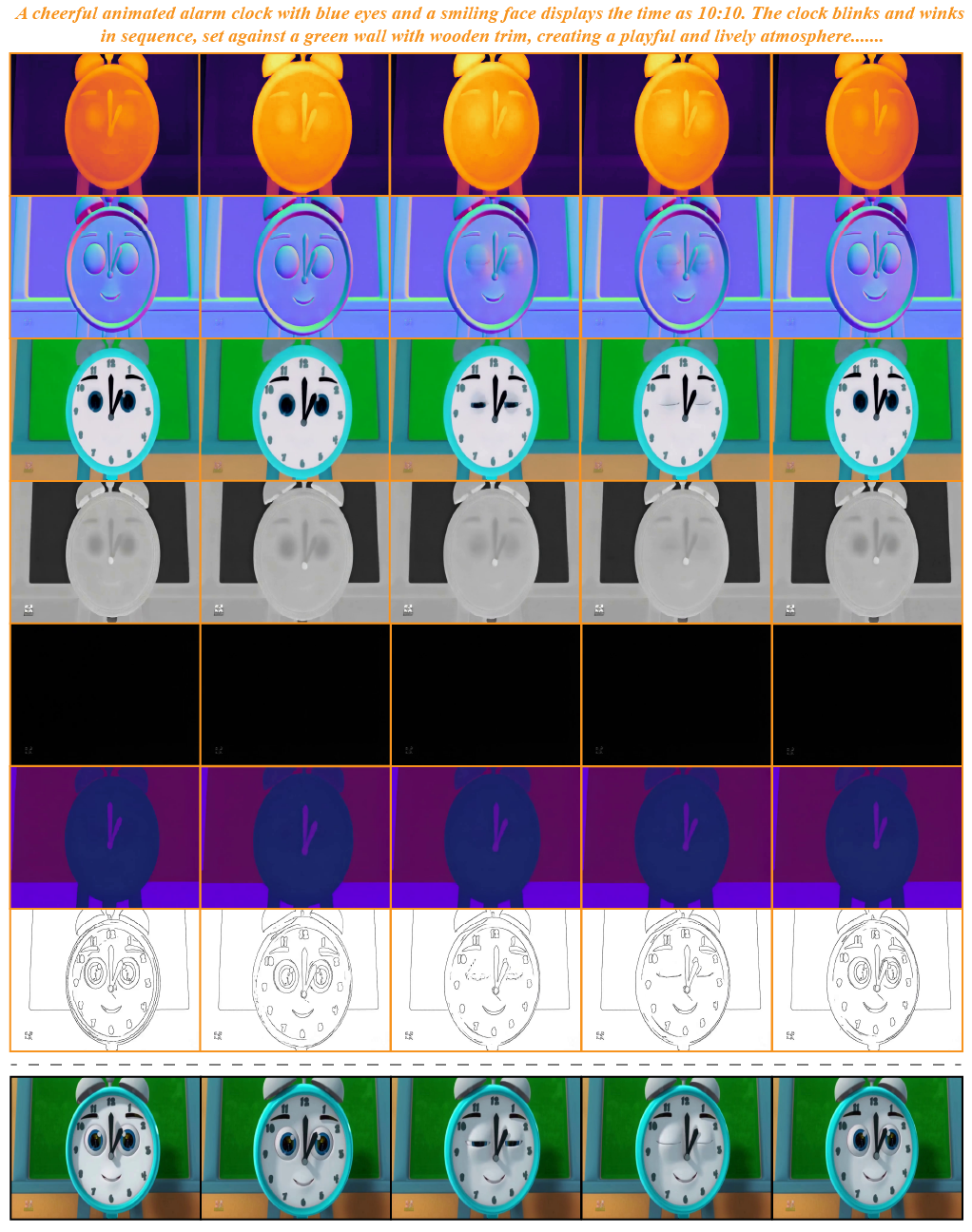}
    \caption{
    \textbf{Additional qualitative results on anime-style video generation.}
    Visualizations of \abbname{} demonstrating its multimodal video generation capability across all modality prediction tasks.
    Our framework maintains stable performance and visual coherence across diverse anime-style scenarios. \textit{Metallic} appears completely black in this visualization, indicating a metallic value close to 0.
    }
  \label{fig:supp_cond_all_2}
\end{figure*}

Figure~\ref{fig:supp_cond_all} shows our generation results when all modalities are used as conditioning inputs. The outputs exhibit high photorealism, as evidenced by physically accurate effects such as reflections on the watch glass. Moreover, we find that our method generalizes well to anime-style scenes, as demonstrated in Figure~\ref{fig:supp_cond_all_2}.

\end{document}